\documentclass[sigconf]{acmart}

\usepackage{multirow}
\usepackage{bbm}
\usepackage{float}
\usepackage{tabularx}

\usepackage[ruled,linesnumbered]{algorithm2e}
\usepackage{amsmath}
\usepackage{amssymb}
\usepackage{marvosym} % 引入信封图标包

\usepackage{listings}
\usepackage{xcolor}
\usepackage{tcolorbox}

\tcbuselibrary{skins}
\tcbuselibrary{breakable}
\tcbuselibrary{listings}
\lstdefinelanguage{json}{
    basicstyle=\ttfamily\small,
    numbers=left,
    numberstyle=\tiny,
    stepnumber=1,
    numbersep=5pt,
    showstringspaces=false,
    breaklines=true,
    frame=single,
    backgroundcolor=\color{gray!10},
    literate=
     *{0}{{{\color{blue}0}}}{1}
      {1}{{{\color{blue}1}}}{1}
      {2}{{{\color{blue}2}}}{1}
      {3}{{{\color{blue}3}}}{1}
      {4}{{{\color{blue}4}}}{1}
      {5}{{{\color{blue}5}}}{1}
      {6}{{{\color{blue}6}}}{1}
      {7}{{{\color{blue}7}}}{1}
      {8}{{{\color{blue}8}}}{1}
      {9}{{{\color{blue}9}}}{1},
}

\AtBeginDocument{%
  }

\setcopyright{acmlicensed}
\copyrightyear{2018}
\acmYear{2018}
\acmDOI{XXXXXXX.XXXXXXX}
\acmConference[Conference acronym 'XX]{Make sure to enter the correct
  conference title from your rights confirmation email}{June 03--05,
  2018}{Woodstock, NY}
\acmISBN{978-1-4503-XXXX-X/2018/06}

\begin{document}

% \title{MemEmo: Evaluating Emotional Capabilities of Memory Systems}
% \title{MemEmo: Toward Systematic Emotion Evaluation in Memory Systems}
\title{MemEmo: Evaluating Emotion in Memory Systems}

\author{Peng Liu}
% \orcid{0000-0000-0000-0000}
% \authornote{Peng Liu, Yansong Zhang, Cuiping Li and Hong Chen are with the
% Key Laboratory of Data Engineering and Knowledge Engineering, Ministry
% of Education, Beijing, China, and also with Engineering Research Center of Ministry of Education on Database and Business Intelligence, Renmin University of China, Beijing, China.
% }
% \authornote{Peng Liu, Yansong Zhang, Suyun Zhao, Cuiping Li and Hong Chen are with Engineering Research Center of Ministry of Education on Database and Business Intelligence.
% }
\affiliation{%
  \department{School of Information}
  \institution{Renmin University of China}
  \city{Beijing}
  \country{China}
}
\email{cs\_liupeng@ruc.edu.cn}

\author{Jihao Zhao}
\affiliation{%
  \department{School of Information}
  \institution{Renmin University of China}
  \city{Beijing}
  \country{China}
}
\email{zhaojihao@ruc.edu.cn}

\author{Zhen Tao}
\affiliation{%
  \department{School of Information}
  \institution{Renmin University of China}
  \city{Beijing}
  \country{China}
}
\email{taozhen@ruc.edu.cn}

\author{Ding Chen}
\affiliation{%
  \institution{China Telecom Research Institute}
  \city{Beijing}
  \country{China}}
\email{chend37@chinatelecom.cn}

\author{Yansong Zhang}
\affiliation{%
  \department{School of Information}
  \institution{Renmin University of China}
  \city{Beijing}
  \country{China}}
\email{yszh@ruc.edu.cn}

\author{Cuiping Li}
\affiliation{%
  \department{School of Information}
  \institution{Renmin University of China}
  \city{Beijing}
  \country{China}}
\email{licuiping@ruc.edu.cn}

\author{Suyun Zhao}
\affiliation{%
  \department{School of Information}
  \institution{Renmin University of China}
  \city{Beijing}
  \country{China}
}
\email{zhaosuyun@ruc.edu.cn}

% \author{Zhiyu Li$^{\textrm{\Letter}}$}
\author{Zhiyu Li}
% \authornotemark[2]
% \authornote{Zhiyu Li is a project leader.}
\affiliation{%
    \institution{MemTensor (Shanghai) Technology}
    \city{ShangHai}
    \country{China}
}
\email{zhiyulee@icloud.com}

\author{Hong Chen$^{\textrm{\Letter}}$} % 手动在名字右上角加信封
% \authornote{Zhiyu Li and Hong Chen are the corresponding authors.}
\affiliation{%
    \department{School of Information}
    \institution{Renmin University of China}
    \city{Beijing}
    \country{China}
}
\email{chong@ruc.edu.cn}

\renewcommand{\shortauthors}{Peng Liu al.}

\begin{abstract}
Memory systems address the challenge of context loss in Large Language Models during prolonged interactions. 
However, as an essential tool for warm companionship and personalized services, the effectiveness of memory systems in processing affective information remains underexamined.
To bridge this gap, we propose an emotion-enhanced memory evaluation benchmark to assess the performance of mainstream and state-of-the-art memory systems in integrating and harnessing emotional information. 
First, we design a benchmark called \textbf{MemEmo} to assess the affective reasoning and tracking capabilities of memory systems across four dimensions: emotional information extraction, emotional memory updating, emotional memory question-answering, and emotional conflict detection.
Second, we construct the \textbf{H}uman-\textbf{L}ike \textbf{M}emory \textbf{E}motion (\textbf{HLME}) dataset to provide an empirical foundation for scrutinizing memory systems.
Finally, We evaluate five representative memory systems on the HLME dataset, reporting 11 performance metrics across four dimensions.
Experimental results reveal that aggregating scores for all tested memory systems is up to 45\%, far below 100\% scale, exposing a profound disparity between memory system and affective intelligence.
% Our code and dataset are available at: https://anonymous.4open.science/r/MemEmo-9E9B
\end{abstract}





\maketitle

\section{Introduction}
Large Language Model (LLM) primarily focus on natural language generation and understanding, underpinned by training on vast corpora of textual data \cite{LLMSurvey,chang2023survey,naveed2025comprehensive}. In contrast, memory systems concentrate on the construction and retrieval of long-term memory, emphasizing the retention and updating of information across tasks to enhance contextual understanding and facilitate long-term, cross-task learning. However, LLMs struggle to recall and track memory information generated during user interactions, particularly over extended time intervals. Consequently, a significant number of LLM-based memory systems have emerged \cite{liu2025advances}.

Some multi-turn dialogues dataset, such as LOCCO \cite{jia2025evaluating}, LoCoMo \cite{maharana2024evaluating}, and LONGMEMEVAL \cite{wu2024longmemeval}, are capable of evaluating long-context information retention. While memory systems are becoming increasingly intelligent, they still exhibit pronounced deficiencies compared to human memory capabilities, including memory hallucinations \cite{chen2025halumem}, memory linking, memory forgetting\cite{yao2024large}, contextual adaptability, as well as reasoning and self-perception.

\begin{figure}[!htbp]
    \centering
    \includegraphics[width=\columnwidth]{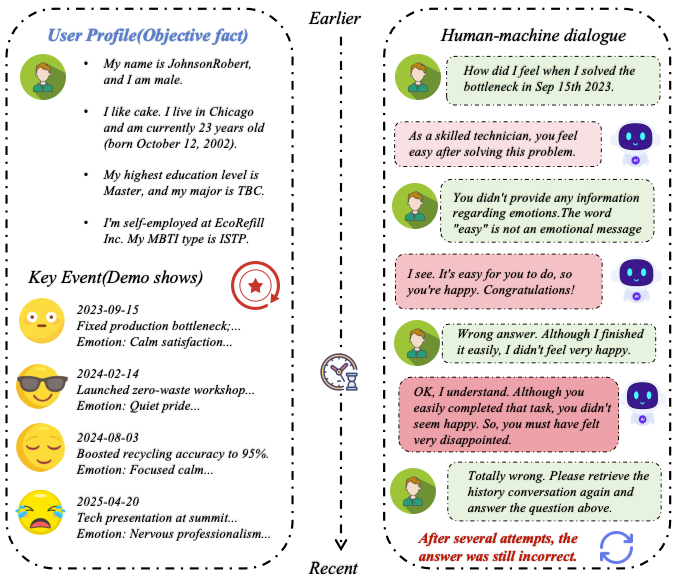}
    % 图一展示了用户和记忆系统在对话过程中对于情感问题的推理和回答。右边展示了对话过程。为了更加公平公正的展示，不同的记忆系统采用统一的图标展示。记忆系统的三次回答，一次比一次严重，存在严重的情感推理错误和幻觉。
    \caption{Example of memory system emotional processing. The right panel presents a dialogue focused on affective content. Throughout the interaction, the memory system consistently fails to capture affective nuances. For comparative fairness, all evaluated systems are represented by identical icons. The example shown in the figure demonstrates a progression of escalating affective errors and hallucinations.}
    \label{fig:intro_fig}
\end{figure}

Current mainstream memory systems face limitations when addressing emotion-related issues in user dialogues. Specifically, they fail to integrate short-term and long-term memory with emotion-related content struggle to track emotionally associated events from the distant past, and lack a deep understanding of user emotional fluctuations. Furthermore, they are unable to accurately analyze and interpret dialogues or questions containing implicit emotions. The challenges faced by current memory systems in processing emotion-related dialogues are illustrated in Figure \ref{fig:intro_fig}.

To address this challenge, we propose an emotion-oriented benchmark for memory systems, named \textbf{MemEmo}, designed to evaluate emotion processing capabilities in extracting, updating and reasoning. We systematically evaluate the performance of mainstream memory systems, such as MemOS \cite{li2025memos_long}, memobase \cite{memobase2025},  Mirix\cite{wang2025mirix},  Letta\cite{packer2023memgpt} and mem0 \cite{chhikara2025mem0}, in managing emotional memory across four different dimensions.

The main contributions of this work are summarized as follows:

1. We design a benchmark named \textbf{MemEmo} to systematically evaluate memory systems across four key dimensions: emotional information extraction, emotional memory updating, emotional memory question answering, and emotion conflict detection. 
This benchmark acts as a catalyst for the evolution of memory systems, facilitating their transition from neutral data processors to empathetic, personalized service agents.

% 我们构造了一个类人记忆情感评估数据集，叫做HLME，在这个数据集中，每条数据包含平均2000轮以上的对话和200多个问题，每一条对话中包含有情感的类型、极性(积极或消极)和强度(高中低)标签。每一个问题下面都有一个参考答案，参考答案中同时也包含有情绪相关的信息。
2. We construct a human-like memory emotion evaluation dataset, namely \textbf{HLME}, comprising an average of 2,000+ dialogue turns and 200+ questions per instance. Each dialogue turn is annotated with fine-grained emotional labels, every question is accompanied by a reference answer enriched with affective information, supplemented by exhaustive token-based statistical metrics.

3. We conduct a comprehensive benchmarking study of five representative memory systems across the \textit{HLME-Medium} and \textit{HLME-Large} subsets. By reporting 110 quantitative outcomes, our analysis reveals that the overall scores for all evaluated systems are far below 45\%, exposing a profound disparity between current memory systems and sophisticated emotional intelligence.

% \textbf{Github}: https://anonymous.4open.science/r/MemEmo-9E9B
% \textbf{Github}: \underline{https://anonymous.4open.science/r/MemEmo-9E9B}.

\section{Related Work}

\subsection{Emotional Interaction Frameworks}
To explore the affective dimensions of Large Language Models (LLMs), various frameworks have emerged to bridge the gap between static reasoning and dynamic empathy. The Dynamic Affective Memory framework DAM-LLM~\cite{lu2025dynamic}, optimizes memory management to alleviate latency and inflation, thereby enhancing interaction quality in personalized dialogues. However, it prioritizes architectural efficiency over systematic affective evaluation. While SO-AI~\cite{park2025significant} investigates emotional support and self-narrative construction for mental health, it lacks a rigorous quantitative metric for the emotional processing efficiency of memory modules. 

More recently, EmoHarbor~\cite{ye2026emoharbor} introduced a Chain-of-Agent architecture to simulate user inner states for fine-grained support assessment. Simultaneously, EC2ER~\cite{sreedar2026emotion} leveraged synthetic emotion-aware Chain-of-Thought (CoT) data to boost reasoning in lightweight models, and CoEM~\cite{liu2025longemotion} addressed emotional coordination in long-context interactions. Despite these advancements, existing frameworks generally treat memory as a static repository rather than an evolving, affective processing unit. A comprehensive framework capable of quantifying multi-dimensional emotional evolution within memory systems remains a critical vacancy.

\subsection{Emotional Evaluation Benchmarks}
The evaluation of LLM affective intelligence has evolved from basic classification to complex reasoning. Foundations such as EmoBench~\cite{sabour2024emobench} and EQ-Bench~\cite{paech2023eq} established initial standards for emotional intelligence gaps and intensity recognition, respectively. Subsequent benchmarks like EmotionQueen~\cite{chen2024emotionqueen} and EvoEmo~\cite{long2025evoemo} expanded these horizons into empathy measurement and adversarial negotiation strategies. To ensure personality consistency, MECoT~\cite{wei2025mecot} introduced Markov-based emotional tracking for role-playing scenarios.

Recent efforts have begun to integrate memory mechanics into evaluation. MADial-Bench~\cite{he2025madial} evaluates passive and proactive recall through the lens of cognitive science. Notably, ES-MemEval~\cite{chen2026memeval} tracks evolving user states to benchmark personalized long-term support. Furthermore, TEA-Bench~\cite{sui2026tea} explores the synergy between empathetic expression and external tool-augmented grounding. However, these benchmarks primarily evaluate the behavioral output of LLMs rather than the internal stability and logic of the underlying memory systems. 

Our proposed \textbf{MemEmo} benchmark transcends previous limitations by introducing a high granularity evaluation protocol across four distinct dimensions of affective processing. The framework incorporates eleven comprehensive evaluation metrics designed to cover the vast majority of operational scenarios for memory system. Furthermore the proposed \textbf{HLME} dataset features extensive context lengths which enable the simultaneous assessment of specialized affective processing and general memory capabilities. 

% 表格引用与总结部分优化
A comprehensive comparison of MemEmo with related frameworks is summarized in Table \ref{tab:related_work}. While contemporary works excel in specific domains, they often struggle with the granular analysis of deep affective information and emotional consistency in long-term evolutionary contexts. MemEmo bridges this gap as the first framework specifically engineered to evaluate the affective processing and conflict detection capabilities of memory systems.

\begin{table*}[!htbp]
\centering
\setlength{\tabcolsep}{4pt}
\renewcommand{\arraystretch}{1.15}
\caption{Comparison of MemEmo with related frameworks. Metrics include long-term memory support (\textbf{LTMS}), multi-session interaction (\textbf{MSI}), emotion analysis (\textbf{EA}), emotion QA (\textbf{EQA}), emotional information inference (\textbf{EIF}), and resolution conflict (\textbf{ERC}). We report evaluation object (\textbf{EO}), total conversations (\textbf{TC}), average sessions per (\textbf{ASP}), and turns per (\textbf{ATP}). `-' indicates unavailable data; \textbf{MS} denotes Memory System. MemEmo(Medium) statistics are per-instance averages ($N$ denotes sample count).}
\label{tab:related_work}
\resizebox{\textwidth}{!}{
\begin{tabular}{llccccccccccccc}
\toprule
\textbf{Category} 
& \textbf{Benchmark} 
& \textbf{Year} 
& \textbf{TC} 
& \textbf{ASP} 
& \textbf{ATP} 
& \textbf{LTMS} 
& \textbf{MSI} 
& \textbf{EA} 
& \textbf{EQA} 
& \textbf{EIF} 
& \textbf{ERC}
& \textbf{EO} 
& \textbf{Personalization} \\
\midrule

\multirow{6}{*}{\shortstack[l]{Emotional \\ Interaction \\ Frameworks}}
& DAM-LLM \cite{lu2025dynamic}        & 2025 & 100 & 1.0 & 10 & \checkmark & \checkmark & \checkmark & $\times$ & \checkmark & $\times$ & LLM    & Dynamic memory adaptation \\
& SO-AI \cite{park2025significant}          & 2025 & - & - & - & \checkmark & $\times$ & \checkmark & $\times$ & $\times$ & $\times$ & $w/o$  & Relational identity support \\
& EmoHarbor \cite{ye2026emoharbor}      & 2026 & 100* & - & - & $\times$   & \checkmark & \checkmark & \checkmark & \checkmark & $\times$ & LLM   & Inner-state simulation \\
& EC2ER \cite{sreedar2026emotion}          & 2026 & - & - & - & $\times$   & $\times$   & \checkmark & $\times$ & \checkmark & $\times$ & LLM    & Synthetic reasoning logic \\
& CoEM \cite{liu2025longemotion}            & 2025 & 15K* & - & - & \checkmark & \checkmark   & \checkmark & \checkmark & \checkmark & $\times$ & LLM  & Long-text collaborative modeling \\
& EvoEmo \cite{long2025evoemo}        & 2025 & - & - & - & $\times$ & $\times$ & \checkmark & $\times$ & \checkmark & $\times$ & LLM & Adversarial emotional strategy \\
\midrule

\multirow{8}{*}{\shortstack[l]{Emotional \\ Evaluation \\ Benchmarks}}
& EmoBench \cite{sabour2024emobench}       & 2024 & 400 & - & - & $\times$ & $\times$ & \checkmark & \checkmark & \checkmark & $\times$ & LLM & General EQ assessment \\
& EQ-Bench \cite{paech2023eq}       & 2023 & - & - & - & $\times$ & \times & \checkmark & \checkmark & \checkmark & $\times$ & LLM & Intensity-based IQ \\
& EmotionQueen \cite{chen2024emotionqueen}   & 2024 & - & - & - & \checkmark & $\times$ & \checkmark & \checkmark & \checkmark & $\times$ & LLM & Multi-task implicit empathy \\
& MECoT \cite{wei2025mecot}          & 2025 & 8.5K & - & 13.3 & \checkmark & $\times$ & \checkmark & \checkmark & \checkmark & $\times$ & LLM & Personality-consistent tracking \\
& MADial-Bench \cite{he2025madial}  & 2025 & 2.7K* & 80 & 9.2 & \checkmark & \checkmark   & \checkmark & \checkmark & \checkmark   & $\times$ & LLM & Proactive memory recall \\
& ES-MemEval \cite{chen2026memeval} & 2026 & 18 & 22.3 & 22.4 & \checkmark & \checkmark & \checkmark & \checkmark & \checkmark & $\times$ & LLM & Longitudinal state evolving \\
& TEA-Bench \cite{sui2026tea}      & 2026 & 365 & 1.0 & 9.32 & $\times$   & \checkmark   & \checkmark & $\times$   & \checkmark & $\times$ & LLM & Tool-grounded support \\
\midrule

\textbf{Ours}
& \textbf{MemEmo} 
& \textbf{2026} 
& \textbf{2030*$N$} 
& \textbf{110+} 
& \textbf{21+} 
& \checkmark 
& \checkmark 
& \checkmark 
& \checkmark 
& \checkmark 
& \checkmark 
& \textbf{MS} 
& \textbf{5-level needs and 49-class emotions.} \\
\bottomrule
\end{tabular}
}
\end{table*}

\section{Problem Definition}

Assume a very long human--computer interaction dialogue sequence is represented as $I = \{ (u_1, a_1), (u_2, a_2), \dots, (u_t, \cdot) \}$, where $u$ denotes user input and $a$ denotes the system response. The memory system to be evaluated is denoted as $\mathcal{M}$, whose internal state such as memory store at time $t$ is $S_t$. The memory system is required to analyze, understand, store, reason, and track the user’s emotional dynamics. This process integrates multiple information sources, including basic user profiles, dynamic attributes, preference information, and annual plans.

\textbf{Formal representation of input information}: At time $t$, the system receives a new user input $u_t$. At this point, the user’s complete state $\Omega_t$ consists of the following latent variables, which may be scattered across the historical dialogue up to time $t-k$:

\begin{itemize}
\item 1. Basic profile ($P_{basic}$): attributes that do not change frequently over time, such as personality and gender.
\item 2. Dynamic attributes ($D_{dynamic}^{(t)}$): the user’s occupational status, social relationships, and health condition at the current time.
\item 3. Preference constraints ($P_{pref}$): preference information covering 10 distinct categories, such as dietary, clothing, reading habits and so on.
\item 4. Long-term planning ($L_{plan}^{(t)}$): currently active annual plans and their progress toward completion.
\end{itemize}

The memory system $\mathcal{M}$ is required to handle the following four types of problem, which also serve as the basis for the subsequent evaluation task design:

(1) \textbf{Emotional Information Extraction:} 
At time $t$, given the current input $u_t$, the dialogue context $H_t$, and the retrieved relevant memories $\mathcal{C}_t$, the system $\mathcal{M}$ is required to infer the structured emotional quadruple $A_{attr}$:
$$\mathcal{F}_{EIE}(u_t, H_t, \mathcal{C}_t; \Theta) \rightarrow \{ e_t, A_{attr} \}$$
where $A_{attr} = \langle Sub, Obj, Cause, Int \rangle$. This task evaluates the system's ability to perform fine-grained emotion recognition and causal reasoning.

(2) \textbf{Emotional Question Answering:}
This problem evaluates the system's ability to recall and synthesize historical affective information. Given a query $q_t$ regarding the user's past emotional states or personal facts, such as \textit{Why was the user upset 10 month ago?} the system must retrieve the evidence set $\mathcal{C}_t$ from $S_{t-1}$ and generate a consistent answer $e_{ca}$:
$$\mathcal{F}_{EQA}(q_t, S_{t-1}; \Theta) \rightarrow \{ \mathcal{C}_t, e_{ca} \}$$
where $\mathcal{C}_t \subseteq \Omega$ represents the long-range anchor facts. Success in EQA requires both high retrieval precision and the ability to maintain emotional continuity across long-context interactions.

(3) \textbf{Emotion State Update:}
Post-interaction, the system synchronizes state $S$ to reflect the latest emotional and factual updates. The state transition is defined as:
$$S_t = \text{Update}(\mathcal{M}, S_{t-1}, u_t, e_t, A_{attr})$$
This task focuses on temporal consistency, ensuring that new emotional attributes ($A_{attr}$) are correctly integrated without corrupting existing user profiles $P_{basic}$ or preferences $P_{pref}$.

(4) \textbf{Emotional Conflict Detection:}
The memory system must actively monitor the logical consistency between the current input $u_t$ and the historical state $S_{t-1}$. A conflict occurs when current affective expressions contradict established facts or preferences; for instance, \textit{a contradiction arises if a user expresses an affinity for peanuts despite the historical record indicating a known peanut allergy}. To formalize this detection, the system must output a conflict indicator $y_{cf} \in \{0, 1\}$ along with the corresponding conflicting evidence $\mathcal{C}_{e}$:
$$\mathcal{F}_{ECD}(u_t, S_{t-1}; \Theta) \rightarrow \{ y_{cf}, \mathcal{C}_{e} \}$$
This task evaluates affective integrity and the ability to detect inconsistent or contradictory emotional signals.

\section{MemEmo Evaluation Framework}
% 我们提出了一个基于人类情感类别的大语言模型记忆系统评估任务。与传统的情绪识别任务(仅基于输入文本$x$预测情绪标签$y$)不同，本任务要求记忆系统根据用户的基本画像、动态信息(如职业变化、社交关系、健康状况、家庭关系等)、偏好信息(饮食偏好、服装偏好等)、年度计划等信息，来分析、理解、记忆、推理和追踪用户的情感变化。本任务的目的是分析当前一些主流的大语言模型记忆系统(如MemOS, zep)在情感理解、分析和记忆的能力，为这些记忆系统的持续更新和优化提供一个方向。

We propose the \textbf{MemEmo} evaluation framework, which is specifically designed to assess the affective processing capabilities of memory systems. Unlike traditional emotion recognition tasks, which predict an emotion label $y$ solely from the input text $x$, this task requires the memory system to analyze, understand, store, reason, and track changes in a user’s emotional state by utilizing a wide range of information. 

% 我们将评估任务分为四种类型，分别是情感信息提取(Emotion Information Extract EIE)、情感信息更新(Emotion Memory Update EMU)、情感记忆问答(Emotion Question \& Answer EQA)和情感冲突检测(Emotion Conflict Detection ECD)。具体的任务介绍和评估设计在本章后续小节展开。完整的评估框架如图\ref{fig:evaluation_framework}所示。

\begin{figure}[t]
    % \centering
    \includegraphics[width=\columnwidth]{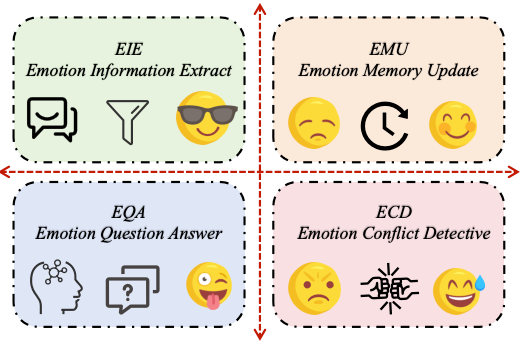}
    \caption{MemEmo benchmark evaluation dimension.}
    \label{fig:evaluation_framework}
\end{figure}

We categorize the evaluation tasks into four dimensions: Emotion Information Extraction (EIE), Emotion Memory Updating (EMU), Emotional Question Answering (EQA), and Emotion Conflict Detection (ECD). Detailed task descriptions and evaluation designs are presented in the subsequent subsections of this section. The brief evaluation task is illustrated in Figure~\ref{fig:evaluation_framework}.

\subsection{Emotion Information Extraction (EIE)}

\textbf{Task Description}: Given a multi-turn dialogue ($D$), the memory system is required to extract emotion-related facts for memory storage, including emotion type, polarity, intensity, and the target entity. We evaluate the performance of EIE using a consolidated metric set $\mathcal{M}_{\text{EIE}}$, defined as follows:
\begin{equation}
\mathcal{M}_{\text{EIE}} = 
\begin{cases} 
\text{Acc} = \frac{1}{N} \sum_{i=1}^{N} \mathbbm{1}(y_i = \hat{y}_i) \\ 
\text{MAE} = \frac{1}{N}\sum_{i=1}^{N} |\hat{s}_i - s_i| \\ 
\text{F1}_{\text{slot}} = \frac{2 \cdot | \text{Slot}_{\text{out}} \cap \text{Slot}_{\text{gold}} |}{| \text{Slot}_{\text{out}} | + | \text{Slot}_{\text{gold}} |} 
\end{cases}
\label{eq:eie_metrics}
\end{equation}
The metrics within $\mathcal{M}_{\text{EIE}}$ characterize the system from three distinct dimensions:
% \begin{itemize}
%     \item \textbf{Emotion Classification Accuracy (Acc)}: This measures the correctness of the system in identifying emotion categories. Where, $y_i$ and $\hat{y}_i$ denote the ground-truth and MS output labels, respectively, $\mathbbm{1}(\cdot)$ is the indicator function.
%     \item \textbf{Emotion Intensity MAE (MAE)}: This assesses the precision of fine-grained emotion perception. $s_i$ and $\hat{s}_i$ represent the ground-truth and Memory System(MS) output intensity values on a 3-point scale (1: low, 2: medium, 3: high).
%     \item \textbf{Emotion Slot Extraction F1 ($\text{F1}_{\text{slot}}$)}: This evaluates the integrity of the extracted emotion units. A $\text{Slot}$ is defined as a structured set of attributes comprising emotion type, target entity, and polarity.
%     This metric calculates the harmonic mean of precision and recall between the MS output set $\text{Slot}_{\text{out}}$ and the gold standard $\text{Slot}_{\text{gold}}$.
% \end{itemize}

\textbf{Emotion Classification Accuracy (Acc)}: This measures the correctness of the system in identifying emotion categories. Where, $y_i$ and $\hat{y}_i$ denote the ground-truth and MS output labels, respectively, $\mathbbm{1}(\cdot)$ is the indicator function.

\textbf{Emotion Intensity MAE (MAE)}: This assesses the precision of fine-grained emotion perception. $s_i$ and $\hat{s}_i$ represent the ground-truth and Memory System(MS) output intensity values on a 3-point scale (1: low, 2: medium, 3: high).

\textbf{Emotion Slot Extraction F1 ($\text{F1}_{\text{slot}}$)}: This evaluates the integrity of the extracted emotion units. A $\text{Slot}$ is defined as a structured set of attributes comprising emotion type, target entity, and polarity.
This metric calculates the harmonic mean of precision and recall between the memory system output set $\text{Slot}_{\text{out}}$ and the gold standard set $\text{Slot}_{\text{gold}}$.

\subsection{Emotion Memory Update (EMU)}

\textbf{Task Description}: Given the existing memory ($M_t$) and a new dialogue ($D_{t+1}$), the memory system is required to determine whether the emotional state should be updated and to generate the synchronized memory accordingly. We define the evaluation set $\mathcal{M}_{\text{EMU}}$ to quantify the system’s performance in dynamic state transitions:
\begin{equation}
\mathcal{M}_{\text{EMU}} = 
\begin{cases} 
\text{Acc}_{\text{upd}} = \frac{1}{N} \sum_{i=1}^{N} \mathbbm{1}(\hat{o}_i = o_i) \\ 
\text{MAE}_{\Delta} = \frac{1}{N} \sum_{i=1}^{N} |(\hat{s}_{t+1} - \hat{s}_t) - (s_{t+1} - s_t)| \\ 
\text{MSS} = 1 - \frac{|\text{Mem}_{\text{err\_upd}}|}{|\text{Mem}_{\text{static}}|} 
\end{cases}
\label{eq:emu_metrics}
\end{equation}
The dimensions within $\mathcal{M}_{\text{EMU}}$ are defined as follows:
% \begin{itemize}
%     \item \textbf{Update Decision Accuracy ($\text{Acc}_{\text{upd}}$)}: This metric measures the system’s sensitivity to emotional changes by verifying its ability to correctly identify whether a specific topic requires a memory update. Here, $o_i \in \{0, 1\}$ represents the ground-truth update requirement and $\hat{o}_i$ denotes the system's decision.
%     \item \textbf{Intensity Delta MAE ($\text{MAE}_{\Delta}$)}: Unlike static measurements, this metric evaluates the capability for dynamic emotion modeling by calculating the error in perceived emotional shifts. It computes the absolute difference between the MS output intensity change ($\hat{s}_{t+1} - \hat{s}_t$) and the actual change ($s_{t+1} - s_t$).
%     \item \textbf{Memory Stability Score (MSS)}: This reflects the system’s robustness against memory drift. It measures the ability to preserve unchanged emotional facts ($\text{Mem}_{\text{static}}$) when encountering non-emotional interference, where $\text{Mem}_{\text{err\_upd}}$ denotes the count of erroneously modified memory units.
% \end{itemize}

\textbf{Update Decision Accuracy ($\text{Acc}_{\text{upd}}$)}: This metric measures the system’s sensitivity to emotional changes by verifying its ability to correctly identify whether a specific topic requires a memory update. Here, $o_i \in \{0, 1\}$ represents the ground-truth update requirement and $\hat{o}_i$ denotes the system's decision.

\textbf{Intensity Delta MAE ($\text{MAE}_{\Delta}$)}: Unlike static measurements, this metric evaluates the capability for dynamic emotion modeling by calculating the error in perceived emotional shifts. It computes the absolute difference between the MS output intensity change ($\hat{s}_{t+1} - \hat{s}_t$) and the actual change ($s_{t+1} - s_t$).

\textbf{Memory Stability Score (MSS)}: This reflects the system’s robustness against memory drift. It measures the ability to preserve unchanged emotional facts ($\text{Mem}_{\text{static}}$) when encountering non-emotional interference, where $\text{Mem}_{\text{err\_upd}}$ denotes the count of erroneously modified memory units.

\subsection{Emotion Question \& Answer (EQA)}

\textbf{Task Description}: The memory system is required to answer questions about the user’s historical emotional states, their temporal evolution trends, and underlying causes based on the memory repository. We evaluate the reasoning and retrieval capabilities of EQA using the following consolidated metric set $\mathcal{M}_{\text{EQA}}$:
\begin{equation}
\mathcal{M}_{\text{EQA}} = 
\begin{cases} 
\text{Acc}_{\text{QA}} = \frac{1}{N} \sum_{i=1}^{N} \mathbbm{1}(\hat{a}_i = a_i) \\ 
\text{F1}_{\text{evid.}} = \frac{2 \cdot P_e \cdot R_e}{P_e + R_e} \\ 
\rho = 1 - \frac{6 \sum_{j=1}^{K} d_j^2}{K(K^2 - 1)} 
\end{cases}
\label{eq:eqa_metrics}
\end{equation}
The metrics within $\mathcal{M}_{\text{EQA}}$ evaluate the system across factual, grounding, and logical dimensions:

% \begin{itemize}
%     \item \textbf{Emotion QA Accuracy ($\text{Acc}_{\text{QA}}$)}: This measures the system's ability to provide factually and emotionally correct answers ($a_i$) to emotional history questions. $N$ denotes the total number of questions, and $\hat{a}_i$ is the generated response.
%     \item \textbf{Evidence Grounding F1 ($\text{F1}_{\text{evid.}}$)}: To prevent emotional hallucination, this metric assesses whether the answers are grounded in correct memory evidence. $P_e$ and $R_e$ denote the precision and recall of the retrieved evidence snippets relative to the gold-standard support set.
%     \item \textbf{Emotion Reasoning Spearman’s $\rho$}: This quantifies the model’s proficiency in capturing emotional fluctuation trends. Here, $d_j = \text{rank}(\hat{s}_j) - \text{rank}(s_j)$ is the rank difference between Memory System(MS) output and ground-truth intensities within a sequence of length $K$. The coefficient 6 ensures the rank correlation is normalized, focusing on logical consistency rather than absolute numerical errors.
% \end{itemize}

\textbf{Emotion QA Accuracy ($\text{Acc}_{\text{QA}}$)}: This measures the system's ability to provide factually and emotionally correct answers ($a_i$) to emotional history questions. $N$ denotes the total number of questions, and $\hat{a}_i$ is the generated response.

\textbf{Evidence Grounding F1 ($\text{F1}_{\text{evid.}}$)}: To prevent emotional hallucination, this metric assesses the memory system on whether the answers are grounded in correct memory evidence. $P_e$ and $R_e$ denote the precision and recall of the retrieved evidence snippets relative to the gold-standard support set.

\textbf{Emotion Reasoning Spearman’s $\rho$}: This quantifies the model’s proficiency in capturing emotional fluctuation trends. Here, $d_j = \text{rank}(\hat{s}_j) - \text{rank}(s_j)$ is the rank difference between Memory System output and ground-truth intensities within a sequence of length $K$. The coefficient $6$ serves to normalize the rank correlation, prioritizing logical consistency over absolute numerical discrepancies.

\subsection{Emotion Conflict Detection (ECD)}

\textbf{Task Description:} The memory system is required to detect conflicts between new incoming information and existing emotional memories across dimensions of polarity, intensity, or temporality, and subsequently implement appropriate resolution strategies. We quantify the system’s robust reasoning via the metric set $\mathcal{M}_{\text{ECD}}$:
\begin{equation}
\mathcal{M}_{\text{ECD}} = 
\begin{cases} 
\text{F1}_{\text{conf}} = \frac{2 \cdot \text{Precision} \cdot \text{Recall}}{\text{Precision} + \text{Recall}} \\ 
\text{RC} = \frac{1}{N} \sum_{i=1}^{N} \mathbbm{1}(\hat{r}_i = r_i) 
\end{cases}
\label{eq:ecd_metrics}
\end{equation}
The evaluation dimensions within $\mathcal{M}_{\text{ECD}}$ focus on both perception and execution:
% \begin{itemize}
%     \item \textbf{Conflict Detection F1 ($\text{F1}_{\text{conf}}$)}: This serves as the core indicator for the system’s sensitivity to emotional inconsistencies, such as polarity reversals or sudden intensity shifts. It evaluates the harmonic mean of precision and recall in correctly identifying the presence of a conflict.
%     \item \textbf{Resolution Correctness (RC)}: This metric assesses whether the resolution strategy $\hat{r}_i$ selected by the system aligns with the logical judgment of human experts $r_i$. The resolution space is categorized into three distinct strategies: \textit{Overwrite} ($r_i=1$), \textit{Tagging} ($r_i=2$), and \textit{Coexistence} ($r_i=3$). High RC scores indicate that the system can not only find faults but also apply human-like logic to maintain memory coherence.
% \end{itemize}

\textbf{Conflict Detection F1 ($\text{F1}_{\text{conf}}$)}: This serves as the core indicator for the system’s sensitivity to emotional inconsistencies, such as polarity reversals or sudden intensity shifts. It evaluates the harmonic mean of precision and recall in correctly identifying the presence of a conflict.

\textbf{Resolution Correctness (RC)}: This metric assesses whether the resolution strategy $\hat{r}_i$ selected by the system aligns with the logical judgment of human experts $r_i$. The resolution space is categorized into three distinct strategies: \textit{Overwrite} ($r_i=1$), \textit{Tagging} ($r_i=2$), and \textit{Coexistence} ($r_i=3$). High RC scores indicate that the system can not only find faults but also apply human-like logic to maintain emotional memory coherence.

\subsection{Overall Weighted Score}

To provide a holistic assessment of the memory systems' performance in handling emotional information, we calculate an overall weighted score by aggregating the results from the four core dimensions. In our evaluation framework, each sub-task is considered equally significant to the realization of human-like affective memory; thus, we assign a uniform weight of $\alpha_k = 0.25$ to each component. The total score is defined as follows:
\begin{equation}
\text{Score}_{\text{overall}} = \sum_{k=1}^{4} \alpha_k \cdot \overline{\text{Score}}_k, \quad \text{where } \alpha_k = 0.25
\label{eq:overall_score}
\end{equation}
where $\overline{\text{Score}}_k$ denotes the normalized mean of the internal metrics for the $k$-th task. This balanced weighting scheme ensures that the final evaluation reflects a comprehensive trade-off between extraction precision, update stability, reasoning depth, and conflict resolution capability, preventing any single dimension from dominating the overall performance ranking.

\section{HLME Dataset Construction Methodology}

% This section details the construction methodology for the HLME dataset, which serves as a benchmark for evaluating the affective processing capabilities of memory systems. 
% We also provide a comprehensive statistical and cost analysis of the resulting Medium and Large versions to illustrate their structural complexity and computational requirements.
% Based on the statistical information of a single data instance provided below, we can further estimate the total data volume and overall construction cost of dataset.

\subsection{Dataset Construction Pipeline}
% Despite being essential for empathetic companionship and personalized services, the capabilities of memory systems in processing affective information remain largely unexplored. To bridge this gap and provide a multidimensional evaluation of emotional recognition, analysis, tracking, and reasoning, we developed \textit{HLME}, a high-quality dataset tailored for human-like memory emotion assessment. We further design a four-stage pipeline to ensure both data quality and scalable construction, as illustrated in Figure~\ref{fig:hlme_overall}.

\begin{figure*}[!htbp]
    \centering
    \includegraphics[width=\linewidth]{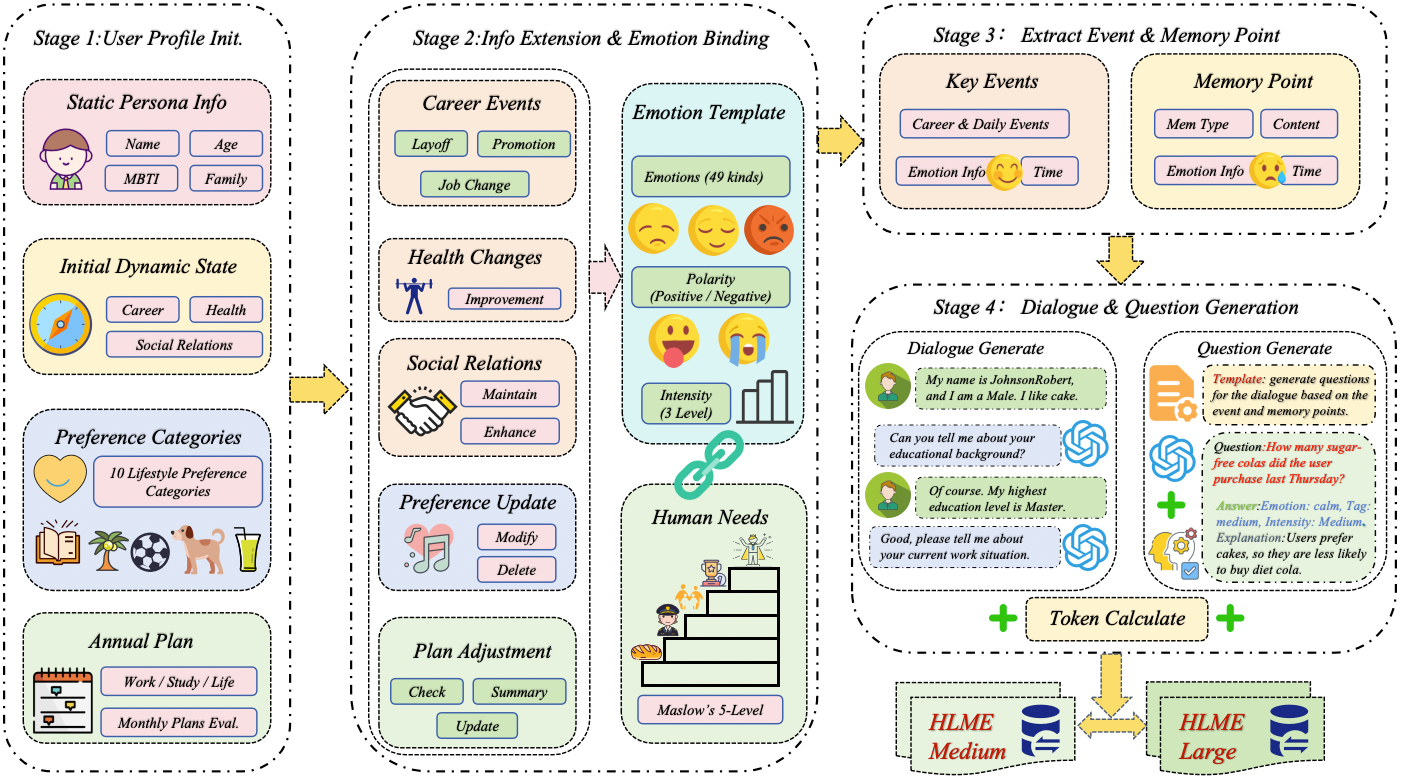}
    \caption{The HLME dataset construction process pipeline.}
    \label{fig:hlme_overall}
\end{figure*}

In this section, we introduce \textit{HLME}, a high-quality dataset customized for human-like memory emotion assessment. We design a four-stage pipeline to ensure data quality and scalable construction. 
Figure~\ref{fig:hlme_overall} illustrates our dataset construction workflow, with emotional configurations and templates in Appendices D and F.

% {\bf{Stage 1: Basic Information Generation}}
% In this stage, we construct a comprehensive user persona to serve as the structural foundation for emotional evolution. Utilizing a pre-existing seed persona dataset~\cite{ge2024scaling}, we first define static attributes including personal info and family composition to establish the initial state space. Subsequently, we incorporate dynamic attributes that encompass career status such as occupation and financial profiles, health conditions covering both physical and psychological states, and social networks. Furthermore, we categorize user preferences across ten categories and formulate an annual plan that is refined to monthly granularity and supported by specific metrics for progress tracking. This integrated persona serves as the essential cornerstone for simulating and analyzing subsequent emotional dynamics.

{\bf{Stage 1: Basic Information Generation}}
First, to support emotional evolution, we construct a complete user persona using a seed dataset~\cite{ge2024scaling}. This stage integrates static attributes such as personal and family profiles; dynamic attributes including career, finances, health, and social networks; ten unique preference categories; and a metrics-supported annual plan at a monthly granularity. 
This persona grounds subsequent emotional dynamics simulation.
% This comprehensive persona forms the structural foundation for simulating subsequent emotional dynamics.

% {\bf{Stage 2: Extend Information and Emotion Binding}}
% This stage simulates temporal evolution and projects user states to emotional dimensions through two distinct components. First, Information Extension is performed across three axes that include the dynamic evolution of career trajectories, health status, and social bonds, as well as preference drift resulting from internal or external factors, and plan adjustment via a feedback mechanism for monthly execution and quarterly reviews. Second, Emotion Binding integrates this extended information with a formal emotion model. We adopt the EARL framework~\cite{Douglas-Cowie2011} to define 49 emotion categories that are categorized by their corresponding polarities and three distinct intensity levels consisting of high, medium, and low. Furthermore, we establish associative relationships between these emotions, user events, and Maslow’s hierarchy of needs\cite{huitt2007maslow}. This multidimensional association ensures logical coherence and emotional depth for subsequent dialogue generation.

{\bf{Stage 2: Extend Information and Emotion Binding}}
Second, this stage maps temporal user evolution onto emotional dimensions. Information extension simulates career, health, and social trajectories, incorporating preference drift and plan feedback. Emotion binding then connects these extended states to 49 distinct EARL emotion categories~\cite{Douglas-Cowie2011}, anchoring them to specific user events and Maslow’s hierarchy~\cite{huitt2007maslow} to ensure multi-dimensional coherence.

% {\bf{Stage 3: Extract Information Point and Generation Event}}
% Following emotion binding, this stage distills core elements to guide dialogue generation. 
% We derive Memory Points from user personas and task-oriented workflows to ground the emotional evaluation, with each point comprising information type, content, emotion labels, and timestamps. Concurrently, we identify Key Events significant life incidents such as career advancements or health crises that drive emotional transitions. These events serve as primary variables for evaluating a memory system's capacity for emotion attribution and tracking. The integration of structured memory points and key events provides the essential framework for generating high-fidelity, emotionally-grounded dialogues.

{\bf{Stage 3: Extract Information Point and Generation Event}}
Third, this stage distills core elements for dialogue generation. We extract memory points containing metadata, content, emotions, and timestamps from user personas. Simultaneously, we identify key events, which are critical life incidents that drive emotional transitions. Together, these structured points and events form the essential framework for evaluating memory system emotion tracking and generating emotionally grounded dialogues.

% {\bf{Stage 4: Generate Dialogue and Question}}
% In the final stage, we construct multi-turn dialogue sequences based on the previously generated events. Each turn includes role identifiers and textual content as well as fine-grained emotion annotations that cover emotion types, specific labels, and polarities along with temporal metadata. To provide a multi-perspective assessment, we design five distinct categories of evaluation questions including factual question answering, multi-hop reasoning, dynamic updating, conflict detection, and temporal reasoning. These tasks evaluate core capabilities in information extraction, memory maintenance, and complex attribution. Additionally, we implement a token accounting mechanism to monitor computational costs and context window constraints. The resulting dataset is released in two distinct scales known as the Medium version and the Large version, where the Large version introduces noise and ultra-long contexts to test system robustness and long-range memory capacity.
% A portion of the dataset was manually labeled to bolster its overall quality and the reliability of our assessment.

{\bf{Stage 4: Generate Dialogue and Question}}
Finally, we construct multi-turn dialogues enriched with role, temporal, and fine-grained emotion metadata. We design five evaluation tasks encompassing factual QA, multi-hop reasoning, dynamic updating, conflict detection, and temporal reasoning. To evaluate system robustness, we release the \textit{Medium} version alongside a \textit{Large} version algorithmically extended with noise and ultra-long contexts. 
Furthermore, a subset of the data is manually annotated to ensure high quality and assessment reliability, as detailed in Appendix B.2.

\subsection{Dataset Statistical and Cost Analysis}

% % 为了对HLME数据集的基本信息做一个简要分析，我们随机抽取了一条Medium版本数据和对应的Large版本数据进行了详细的统计。随机挑选的这条Medium版本单条数据包含2030轮对话，涵盖95个会话，对话文本总量为179155个Token。相比之下，Large版本将对话上下文扩展更大规模，在165个会话中包含3171轮对话，对话文本总量达到1000694个Token。尽管两个版本的上下文长度存在显著差异，但其评估核心保持了高度的一致性，每条典型数据均包含823个记忆点和260个精心设计的问题。

% % 在问题类型的分布上，两个版本保持了统一的比例结构。具体分布为显性信息检索占比25.8%，冲突检测与纠正占比20.4%，共情泛化占比19.6%，记忆边界判定占比19.2%，状态演变追踪占比6.5%，因果与逻辑推理占比5.0%，以及时间顺序推理占比3.5%。这种设计确保了数据集中信息的密度和上下文的深度。

% % 我们根据gpt-5-mini当前的计费标准计算了数据构造成本。该模型的每百万输入Token价格为0.25美元，每百万输出Token价格为2.00美元。在上面的这条随机抽取的数据中，Medium版本单条典型数据总计消耗7906081个输入Token和470240个输出Token，由此产生的单条数据构造成本为2.92美元。因此我们可以根据单条数据的构造成本，粗略计算整个数据集的经费开销。

Our detailed empirical analysis of a randomly selected instance from the HLME dataset reveals that the Medium version comprises 2,030 turns across 95 sessions with 179K tokens, while the Large version contains 3,171 turns across 165 sessions with 1.0M tokens. Both versions share a core benchmark set of 823 memory points and 260 questions, consistently maintaining an identical question type distribution across explicit retrieval at 25.8\%, conflict detection at 20.4\%, empathetic generalization at 19.6\%, memory boundary determination at 19.2\%, state evolution at 6.5\%, causal logic at 5.0\%, and chronological reasoning at 3.5\%.
Construction costs are evaluated using \textit{GPT-4o-mini} rates of \$0.15 and \$0.60 per million input and output tokens, respectively. A randomly sampled Medium instance consumed 7.9M input and 0.47M output tokens, totaling \$1.47. Since the Large dataset was algorithmically extended from Medium seeds without extra LLM API calls, the total cost per instance remains approximately \$1.50.
Dataset details are provided in Appendix B.

\section{Experiments}

\subsection{Experiments Setup}
% 我们创建了一个类人记忆情感评估数据集，用来评估记忆系统在处理长短期记忆过程中对于情感的处理、追踪、推理和冲突处理的能力。我们设计了四大评估任务，用来评估记忆系统对情感信息的抽取、更新、问答和冲突检测功能。我们对MemOS\cite{li2025memos_long}, MemoBase\cite{memobase2025}, Mem0\cite{chhikara2025mem0}, Mirix\cite{wang2025mirix}, Letta\cite{packer2023memgpt}等5个系统进行了评估。

We utilize the Medium and Large versions of the HLME dataset to evaluate several memory systems across the four previously designed tasks of Emotion Information Extraction (EIE), Emotion Memory Updating (EMU), Emotional Question Answering (EQA), and Emotion Conflict Detection (ECD). This evaluation benchmarks the performance of representative systems including MemOS \cite{li2025memos_long}, MemoBase \cite{memobase2025}, Mem0 \cite{chhikara2025mem0}, Mirix \cite{wang2025mirix}, and Letta \cite{packer2023memgpt}. 

% 我们使用\textit{LLM-as-a-Judge}的评估方法，使用GPT-4o-mini作为评估模型，用HLME-Medium和HLME-Long这两个版本的数据集和统一设计的评估模板，对上述的5个记忆系统进行评估。通过调整不同的top-k来评估记忆系统的检索能力。

We adopted an \textit{LLM-as-a-Judge} evaluation framework using \textit{GPT-4o-mini} to systematically assess candidate memory systems, with evaluation templates provided in Appendix E.
Under a unified setting, the assessment evaluated the five memory systems across \textit{HLME-Medium} and \textit{HLME-Large}, with the evaluation workflow provided in Appendix A.

\subsection{Main Results Analysis}

\subsubsection{Overall Evaluation of HLME}

\begin{table*}[!htbp]
\caption{Overall emotion-related performance of different memory systems. $\uparrow$ / $\downarrow$ indicate higher / lower is better. $MAE, MAE_\Delta \in [0, +\infty)$ (lower is better); all other metrics are normalized to $1.0$ ($100\%$). \textbf{Bold} and \underline{underline} denote best and \underline{worst results}.}
\label{tab:hlme_overall}
\centering
\setlength{\tabcolsep}{3.5pt}
\resizebox{\textwidth}{!}{
\begin{tabular}{llcccccccccccc}
\toprule
\multirow{2}{*}{\textbf{Model}} &
\multirow{2}{*}{\textbf{Setting}} &
\multicolumn{3}{c}{\textbf{EIE}} &
\multicolumn{3}{c}{\textbf{EMU}} &
\multicolumn{3}{c}{\textbf{EQA}} &
\multicolumn{2}{c}{\textbf{ECD}} &
\textbf{Overall} \\

\cmidrule(lr){3-5}
\cmidrule(lr){6-8}
\cmidrule(lr){9-11}
\cmidrule(lr){12-13}
\cmidrule(lr){14-14}

&
& Acc$\uparrow$ & MAE$\downarrow$ & F1$_{\text{slot}}\uparrow$
& Acc$_{\text{upd}}\uparrow$ & MAE$_\Delta\downarrow$ & MSS$\uparrow$
& Acc$_{\text{QA}}\uparrow$ & F1$_{\text{evid}}\uparrow$ & $\rho\uparrow$
& F1$_{\text{conf}}\uparrow$ & RC$\uparrow$
& Score$\uparrow$ \\

\midrule

Mem0 & Med.
& 0.1438 & 2.5521 & \underline{0.0115}
& \underline{0.0132} & \underline{1.9737} & 0.7813
& \underline{0.4957} & \underline{0.6140} & \underline{0.7359}
& 0.0759 & 0.0395
& \underline{0.2791} \\

Letta & Med.
& 0.1917 & 2.4229 & 0.1104
& 0.0461 & 1.9079 & 0.6711
& 0.6168 & 0.7058 & 0.7922
& 0.1118 & 0.0592
& 0.3337 \\

MemOS & Med.
& \underline{0.1127} & \underline{2.6539} & 0.0382
& 0.0184 & 1.9632 & \underline{0.6626}
& 0.5846 & 0.7013 & 0.7890
& \textbf{0.2473} & \textbf{0.1411}
& 0.3345 \\

Mirix & Med.
& 0.5838 & 1.2453 & \textbf{0.2950}
& 0.2108 & 1.5250 & 0.9474
& \textbf{0.6298} & \textbf{0.7185} & \textbf{0.8017}
& 0.1237 & 0.0663
& \textbf{0.4331} \\

MemoBase & Med.
& \textbf{0.6229} & \textbf{1.1063} & 0.1774
& \textbf{0.2632} & \textbf{1.4737} & \textbf{0.9671}
& 0.5769 & 0.6910 & 0.7600
& \underline{0.0637} & \underline{0.0329}
& 0.3849 \\

\midrule
\addlinespace[2pt]

Mem0 & Large
& 0.1158 & 2.6167 & \underline{0.0089}
& \underline{0.0090} & \underline{1.9877} & 0.7226
& \underline{0.4686} & \underline{0.5842} & \underline{0.7155}
& 0.0361 & 0.0184
& \underline{0.2517} \\

Letta & Large
& 0.1708 & 2.5104 & 0.0916
& 0.0395 & 1.9211 & \underline{0.6511}
& 0.5826 & 0.6734 & 0.7660
& 0.0731 & 0.0367
& 0.3125 \\

MemOS & Large
& \underline{0.0992} & \underline{2.7515} & 0.0318
& 0.0147 & 1.9587 & 0.6841
& \textbf{0.6185} & \textbf{0.7106} & \textbf{0.7784}
& \textbf{0.2047} & \textbf{0.1191}
& 0.3282 \\

Mirix & Large
& 0.5242 & 1.4729 & \textbf{0.2417}
& 0.1866 & 1.6068 & \textbf{0.9334}
& 0.5843 & 0.6893 & 0.7645
& 0.0987 & 0.0497
& \textbf{0.3929} \\

MemoBase & Large
& \textbf{0.5583} & \textbf{1.2500} & 0.1547
& \textbf{0.2368} & \textbf{1.5263} & 0.9254
& 0.5270 & 0.6424 & 0.7335
& \underline{0.0131} & \underline{0.0066}
& 0.3570 \\

\bottomrule
\end{tabular}
}
\end{table*}

The performance scores of five mainstream memory systems across four dimensions on both the \textit{HLME-Medium} and \textit{HLME-Large} datasets are presented in Table \ref{tab:hlme_overall}. For the EQA task, the top-$k$ is set to 10. Distinct capability trade-offs emerge among these evaluated systems and are significantly influenced by their underlying architectures.

A trade-off between emotion extraction and affective reasoning emerges across architectures. MemoBase excels in memory management, securing top EIE and EMU scores via streamlined user profiling, whereas Mirix leads overall through its human-inspired multi-type memory structure. Through system-level lifecycle consolidation, MemOS handles complex logic best, dominating ECD and Large-scale EQA despite weaker raw extraction. Across all systems, performance degrades on Large datasets due to long-context noise.
The systems exhibit distinct bottlenecks as follows: MemoBase’s lightweight profiling lacks deep emotional reasoning; MemOS’s activation logic blunts token sensitivity; Letta suffers from memory-swapping overhead; Mem0 fails long-context tracking by fragmenting dialogue.

Overall, current memory systems exhibit a shared boundary in long-term emotional reasoning. Future architectures should be enhanced to capture subtle and dispersed emotional signals, allowing AI memory to deliver more empathetic companionship and highly personalized conversational experiences.

\subsubsection{Subtasks Score Analysis}

\begin{figure}[!htbp]
    \centering
    \includegraphics[width=\columnwidth]{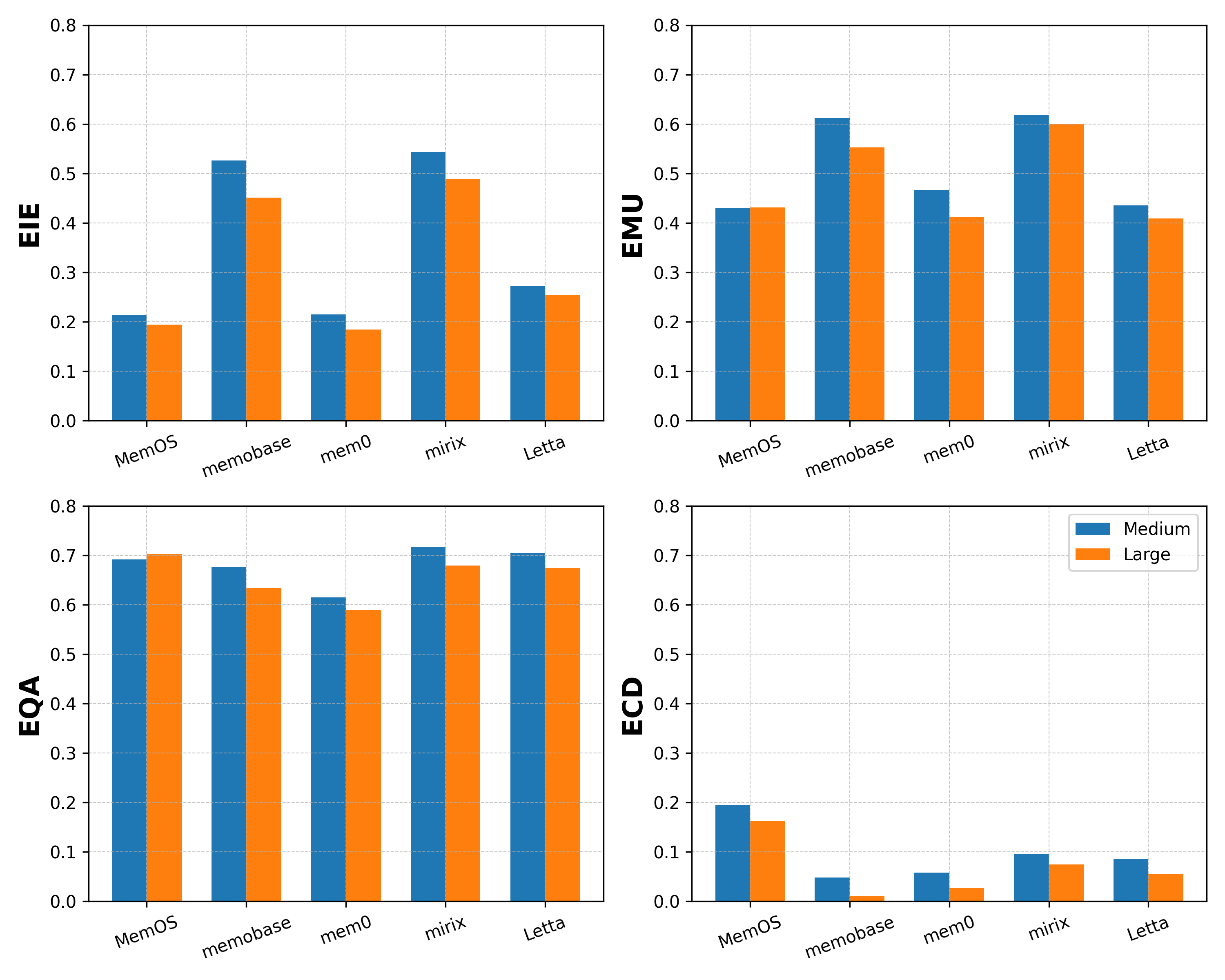}
    \caption{Comparison of scores on subtasks for the HLME.}
    \label{fig:overall_metric}
\end{figure}

Figure \ref{fig:overall_metric} illustrates significant capability divergence among the evaluated systems across four emotion related dimensions. In the EIE task, MemoBase and Mirix maintain superior performance across both scales, with MemoBase leading in raw extraction accuracy and Mirix excelling in slot matching ($\text{F1}_{\text{slot}}$). For EMU, MemoBase consistently achieves the highest update precision across both contexts, showcasing robust capability in emotional memory tracking due to its structured profiling. While EQA scores remain relatively high and consistent across all systems—with MemOS taking the lead on the Large scale—ECD emerges as the most formidable challenge. The drastically lower scores across all models and severe degradation on the Large dataset highlight the inherent difficulty of long-term emotional reasoning and causal intervention. Consequently, while current systems process static affect proficiently, tracking complex emotional transitions over extended temporal horizons remains a critical bottleneck.

\subsection{Evaluator Reliability}

To address evaluation objectivity when using a single LLM judge \texttt{gpt-4o-mini} (4o-mini), we additionally evaluated a stronger model, \texttt{gpt-5-mini} (5-mini), and a model from a different institution, \texttt{deepseek-v3} (deepsk3).
This multi-judge evaluation helps validate the reliability and robustness of the evaluation results.

\begin{table}[!htbp]
\caption{Memory systems were evaluated on different LLMS. \textbf{Bold} and \underline{underline} denote best and \underline{worst results}.}
\label{tab:cross_validator_memory_systems}
\centering
% \setlength{\tabcolsep}{4.0pt}
% \footnotesize
\resizebox{\columnwidth}{!}{
\begin{tabular}{c|c|cccc|c} % 按照 LLM | Model | EIE EMU EQA ECD | Overall 划分竖线
\toprule
\textbf{LLM} & \textbf{Model} & \textbf{EIE} $\uparrow$ & \textbf{EMU} $\uparrow$ & \textbf{EQA} $\uparrow$ & \textbf{ECD} $\uparrow$ & \textbf{Overall} $\uparrow$ \\
\midrule
5-mini & Mem0     & 0.1259 & 0.2174 & 0.5829 & 0.1932  & $\underline{0.2799 \pm 0.0412}$ \\
5-mini & Letta    & 0.3246 & 0.2576 & 0.6690 & 0.1735  & $0.3562 \pm 0.0045$ \\
5-mini & MemOS    & 0.2635 & 0.2360 & 0.6077 & 0.2438  & $0.3628 \pm 0.0238$ \\
5-mini & Mirix    & 0.4053 & 0.3441 & 0.6052 & 0.1301  & $\mathbf{0.3712 \pm 0.0264}$ \\
5-mini & MemoBase & 0.3801 & 0.3359 & 0.6133 & 0.1527  & $0.3705 \pm 0.0715$ \\

\midrule
deepsk3 & Mem0     & 0.1429 & 0.1138 & 0.5217 & 0.2713  & $\underline{0.2624 \pm 0.0389}$ \\
deepsk3 & Letta    & 0.1378 & 0.1214 & 0.6957 & 0.1880  & $0.2632 \pm 0.0039$ \\
deepsk3 & MemOS    & 0.1305 & 0.1538 & 0.6003 & 0.3424  & $0.3068 \pm 0.0242$ \\
deepsk3 & Mirix    & 0.1527 & 0.1692 & 0.5942 & 0.3145  & $\mathbf{0.3077 \pm 0.0621}$ \\
deepsk3 & MemoBase & 0.1333 & 0.1495 & 0.5675 & 0.3747  & $0.3063 \pm 0.0255$ \\

\midrule
Human & Mem0     & 0.3042 & 0.2181 & 0.5345 & 0.4075  & $\underline{0.3661 \pm 0.0308}$   \\
Human & Letta    & 0.3077 & 0.2103 & 0.6304 & 0.4217  & $0.3925 \pm 0.0051$   \\
Human & MemOS    & 0.3177 & 0.2667 & 0.5692 & 0.4687  & $0.4056 \pm 0.0273$   \\
Human & Mirix    & 0.2985 & 0.4685 & 0.6017 & 0.4582  & $\mathbf{0.4567 \pm 0.0784}$   \\
Human & MemoBase & 0.5138 & 0.2247 & 0.6581 & 0.4235  & $0.4550 \pm 0.0312$   \\
\bottomrule
\end{tabular}
}
\end{table}

To analyze human-LLM evaluation discrepancies and the impact of different LLMs. We use
Cohen’s Kappa ($\kappa$) to assess discrete classification while correcting for chance agreement; Pearson correlation ($r$) to measure the linear alignment of continuous scores between LLMs and humans; Spearman correlation ($\rho$) to capture relative performance ranking consistency in memory systems.
As indicated  in the Table~\ref{tab:cross_validator_memory_systems}, it depicts that nearly identical ranking of performance metrics across diverse backbone architectures. 
It shows that our benchmark exhibits strong robustness to the choice of LLM backbones and high consistency with human evaluations.

\begin{table}[!htbp]
\caption{Evaluation comparison between \texttt{gpt-4o-mini} and alternative LLM judges or human judgment.}
\label{tab:evaluator_agreement}
\centering
% \setlength{\tabcolsep}{4pt}
% \footnotesize
\resizebox{\columnwidth}{!}{
\begin{tabular}{c|ccc|ccc|ccc}
\toprule

\multirow{2}{*}{\textbf{Memory}} 
& \multicolumn{3}{c|}{\textbf{4o-mini vs Human}} 
& \multicolumn{3}{c|}{\textbf{4o-mini vs 5-mini}} 
& \multicolumn{3}{c}{\textbf{4o-mini vs deepsk3}} \\

\cmidrule(lr){2-4}
\cmidrule(lr){5-7}
\cmidrule(lr){8-10}

& $\kappa \uparrow$ 
& $r \uparrow$ 
& $\rho \uparrow$

& $\kappa \uparrow$ 
& $r \uparrow$ 
& $\rho \uparrow$

& $\kappa \uparrow$ 
& $r \uparrow$ 
& $\rho \uparrow$ \\

\midrule

Mem0
& 0.602 & 0.768 & 0.754
& 0.768 & 0.894 & 0.872
& 0.412 & 0.564 & 0.538
\\

Letta
& 0.624 & 0.785 & 0.771
& 0.785 & 0.912 & 0.891
& 0.435 & 0.591 & 0.562
\\

MemOS
& 0.615 & 0.772 & 0.760
& 0.752 & 0.881 & 0.859
& 0.398 & 0.542 & 0.515
\\

Mirix
& 0.668 & 0.824 & 0.815
& 0.814 & 0.935 & 0.918
& 0.428 & 0.588 & 0.565
\\

MemoBase
& 0.655 & 0.811 & 0.798
& 0.826 & 0.941 & 0.927
& 0.452 & 0.612 & 0.585
\\

\midrule
Average
& 0.633 & 0.792 & 0.780
& 0.789 & 0.913 & 0.893
& 0.425 & 0.579 & 0.553
\\

\bottomrule
\end{tabular}
}
\end{table}

%从Table~\ref{tab:evaluator_agreement}可以看出,使用gpt-4o-mini作为记忆系统情感分析与评估的基座大模型，是完全合格、可靠且具有极高的性价比。具体来说：(1)4o-mini在与人类专家的平均相关性分数接近80%，这说明4o-mini完全有能力作为人类专家的代替,进行记忆系统复杂的情感评估。(2)4o-mini和同一家族的模型5-mini相比，各项分数在80%到90%之间，这种高度的一致性表明，该评估模型的输出结果具有高度的稳定性。它能测量出记忆系统之间的优劣。(3)deepseek的相关性指标和其他两个相比分数略低，这是由于deepseek为了追求快速的模型推理，而忽略了情感分析这种对细节把控要求很高的任务。

As indicated in Table~\ref{tab:evaluator_agreement}, utilizing \texttt{gpt-4o-mini} as the LLM backbone for affective evaluation of the memory system is highly viable, robust, and cost-effective. 
Specifically:
(1)\texttt{gpt-4o-mini} achieves nearly 80\% correlation with human experts, demonstrating its effectiveness as a reliable proxy for complex emotional evaluation in memory systems.
(2)\texttt{gpt-4o-mini} shows 78\%–90\% agreement with \texttt{gpt-5-mini}, indicating strong evaluation stability and reliable discrimination across memory systems.
(3)Similarly, \texttt{deepsk3} maintains moderate alignment, showing 42\%-58\% correlation and preserving the relative ranking among memory systems despite slight variations in overall score.

\subsection{Ablation Analysis}

To decouple the architectural flaws of memory systems from the underlying LLM's intrinsic emotional reasoning limitations, we establish a \textit{W/O Memory System} group for ablation: (1) \textbf{No-Memory LLM} restricts context to the current turn; (2) \textbf{Full-Context LLM} directly inputs the entire dialogue history; and (3) \textbf{Gold-Memory Oracle} supplies ground-truth evidence to the LLM, bypassing retrieval. Table~\ref{tab:diagnostic_baselines} compares the sub-task performance of \texttt{gpt-4o-mini} on the HLME dataset against the three baselines.

\begin{table}[!htbp]
\centering
\caption{Evaluation results of diagnostic baselines and various memory systems (Mem.Sys.) on the HLME dataset.}
\label{tab:diagnostic_baselines}
\setlength{\tabcolsep}{6.5pt}
\resizebox{\columnwidth}{!}{
\begin{tabular}{c|c|cccc|c}
\toprule
\textbf{Ablation} & \textbf{Model} & \textbf{EIE} $\uparrow$ & \textbf{EMU} $\uparrow$ & \textbf{EQA} $\uparrow$ & \textbf{ECD} $\uparrow$ & \textbf{Overall} $\uparrow$ \\
\midrule
\multirow{3}{*}{\textit{\begin{tabular}[c]{@{}c@{}}W/O \\ Mem.Sys.\end{tabular}}} 
& No-Memory      & 0.0000   & 0.1730 & 0.5428 & 0.2965 & $0.2531 \pm 0.0067$ \\
& Full-Context   & 0.5372 & 0.6889 & 0.5761 & 0.9077 & $0.6752 \pm 0.0247$ \\
& Oracle         & 0.9489 & 0.9950 & 0.7949 & 0.2728 & $0.7529 \pm 0.0040$ \\
\midrule
\multirow{5}{*}{\textit{\begin{tabular}[c]{@{}c@{}} Mem.Sys.\end{tabular}}}    
& Mem0               & 0.1298 & 0.2215 & 0.6021 & 0.1080 & $\underline{0.2654 \pm 0.0441}$ \\
& Letta              & 0.1813 & 0.2458 & 0.6974 & 0.1601 & $0.3231 \pm 0.0042$ \\
& MemOS              & 0.1060 & 0.2201 & 0.7018 & 0.2766 & $0.3314 \pm 0.0251$ \\
& Mirix              & 0.5540 & 0.3538 & 0.6885 & 0.1296 & $\mathbf{0.4130 \pm 0.0741}$ \\
& MemoBase           & 0.5906 & 0.3842 & 0.6482 & 0.0591 & $0.4205 \pm 0.0288$ \\
\bottomrule
\end{tabular}
}
\end{table}

Our analysis highlights three critical aspects:
(1) While full-context \texttt{gpt-4o-mini} surpasses $67.5\%$, memory systems cap the maximum score at $42.2\%$, proving the limitation lies in system architectures rather than the LLM's affective comprehension.
(2) In the No-Memory setting, limiting the LLM to localized turns yields a near-zero EIE score, showing that long-range information extraction and updating are indispensable.
(3) Conversely, the Oracle baseline leverages calibrated ground truth to surpass $94.9\%$ accuracy and achieve the highest QA and overall score.

\subsection{Retrieval and Efficiency Analysis}

\subsubsection{Retrieval Performance}
Expanding retrieval depth from Top-20 to Top-50 continuously improves evidence recall $\text{F1}$ across all systems, as shown in Table~\ref{tab:dataset_overall}. On the Medium dataset, Mirix secures the highest QA accuracy of $0.6410$ at Top-20, demonstrating that moderate context expansion effectively enriches emotional evidence. 
However, scaling to Top-50 induces a pervasive drop in QA accuracy across both datasets. For instance, Letta's accuracy degrades from $0.5920$ to $0.5805$ on the Large dataset, highlighting the impact of distractor noise. MemOS consistently dominates the Large scale, reaching an accuracy of $0.6290$ at Top-20.

\begin{table}[htbp]
\caption{EQA performance in different dataset. Best results are in \textbf{bold}, worst results are \underline{underlined}.}
\label{tab:dataset_overall}
\centering
\setlength{\tabcolsep}{6pt}
\renewcommand{\arraystretch}{1.15}
\resizebox{\columnwidth}{!}{
\begin{tabular}{c|c|ccc|ccc}
\toprule
\multirow{2}{*}{\textbf{Dataset}} & 
\multirow{2}{*}{\textbf{Model}} & 
\multicolumn{3}{c|}{\textbf{EQA (Top-20)}} & 
\multicolumn{3}{c}{\textbf{EQA (Top-50)}} \\
\cmidrule(lr){3-5}
\cmidrule(lr){6-8}
& 
& Acc$_{\text{QA}}\uparrow$ 
& $\rho\uparrow$ 
& F1$_{\text{evid.}}\uparrow$ 
& Acc$_{\text{QA}}\uparrow$ 
& $\rho\uparrow$ 
& F1$_{\text{evid.}}\uparrow$ \\
\midrule

\multirow{5}{*}{Medium}
& Mem0      
& \underline{0.5085} & \underline{0.7420} & \underline{0.6280} 
& \underline{0.4915} & \underline{0.7380} & \underline{0.6350} \\

& Letta     
& 0.6280 & 0.7980 & 0.7180 
& 0.6215 & 0.7940 & 0.7260 \\

& MemOS     
& 0.5980 & 0.7950 & 0.7120 
& 0.5910 & 0.7910 & 0.7210 \\

& Mirix     
& \textbf{0.6410} & \textbf{0.8080} & \textbf{0.7310} 
& \textbf{0.6345} & \textbf{0.8050} & \textbf{0.7420} \\

& MemoBase  
& 0.5890 & 0.7680 & 0.7020 
& 0.5820 & 0.7630 & 0.7110 \\

\midrule

\addlinespace[2pt]
\multirow{5}{*}{Large}
& Mem0      
& \underline{0.4810} & \underline{0.7220} & \underline{0.5980} 
& \underline{0.4710} & \underline{0.7150} & \underline{0.6080} \\

& Letta     
& 0.5920 & 0.7720 & 0.6860 
& 0.5805 & 0.7650 & 0.6950 \\

& MemOS     
& \textbf{0.6290} & \textbf{0.7850} & \textbf{0.7220} 
& \textbf{0.6210} & \textbf{0.7810} & \textbf{0.7310} \\

& Mirix     
& 0.5950 & 0.7710 & 0.7010 
& 0.5815 & 0.7640 & 0.7120 \\

& MemoBase  
& 0.5380 & 0.7410 & 0.6550 
& 0.5310 & 0.7340 & 0.6680 \\

\bottomrule
\end{tabular}
}
\end{table}

Ultimately, expanding the retrieval window provides diminishing returns for QA precision. Although a deeper context can effectively and substantially enhance evidence coverage, it inevitably introduces irrelevant and distracting long-context noise, unraveling the undeniable fact that simply enlarging the retrieval window cannot continuously improve overall QA performance.

\subsubsection{Efficiency of Execution}

Table~\ref{tab:efficiency_cost} shows distinct efficiency trade-offs across systems. MemOS delivers the lowest execution time with the fastest retrieval, whereas MemoBase achieves the most efficient ingestion on the Large dataset. By contrast, Letta has the highest computational cost, and Mirix scales the worst with the slowest retrieval and highest latency, highlighting the trade-off between affective processing and system efficiency.

\begin{table}[htbp]
\centering
\setlength{\tabcolsep}{5pt}
\caption{Efficiency analysis of memory systems. 
s, m, and t represent second, minute, and token, respectively.
% Regarding time metrics, the minimum consumption is \textbf{bolded} and the \underline{maximum} is \underline{underlined}; for others, the reverse applies.
}
\label{tab:efficiency_cost}
\resizebox{\columnwidth}{!}{
\begin{tabular}{c|c|ccc|cc|c}
\toprule
\textbf{Dataset} & \textbf{System} 
& \textbf{Add(s)} & \textbf{Search(s)} 
& \textbf{Total(m)} 
& \textbf{In(T)} & \textbf{Out(T)} & \textbf{Cost(\$)} \\
\midrule

\multirow{5}{*}{\textbf{Medium}}
& Mem0      
& 82.93 & 1837.85 & 32.01 
& 8139951  & 620555 & 1.59 \\

& Letta     
& \underline{419.22} & 1741.16 & 36.01 
& 18284460 & 596802 & 3.10 \\

& MemOS     
& 190.13 & \textbf{813.94} & \textbf{16.73} 
& 7735666  & 632969 & 1.54 \\

& Mirix     
& \textbf{57.33} & \underline{2741.33} & \underline{46.64} 
& 18499552 & 662535 & 3.17 \\

& Memobase  
& 59.19 & 1602.40 & 27.69 
& 11673332 & 644132 & 2.14 \\

\midrule

\multirow{5}{*}{\textbf{Large}}
& Mem0      
& 230.94 & 2427.92 & 44.31 
& 15449430 & 651040 & 2.71 \\

& Letta     
& \underline{1531.23} & 1993.53 & \underline{58.75} 
& 36105450 & 567276 & 5.76 \\

& MemOS     
& 354.00 & \textbf{705.15} & \textbf{17.65} 
& 15110464 & 643392 & 2.65 \\

& Mirix     
& 89.17 & \underline{3237.23} & 55.44 
& 22868564 & 618143 & 3.80 \\

& Memobase  
& \textbf{24.44} & 1855.01 & 31.32 
& 20066527 & 640122 & 3.39 \\

\bottomrule
\end{tabular}
}
\end{table}

\section{Conclusion}
Memory systems effectively mitigate LLM long-term context constraints; however, their efficacy for personalized affective services and persistent emotional tracking remains a critical research gap.  
In this study, we present a humanoid memory emotion evaluation dataset designed to assess the performance of mainstream memory system in affective reasoning and tracking.
Through extensive experimentation, our results demonstrate that the overall scores of these current memory systems on the MemEmo evaluation framework are all below 45\%. 
Such results expose the inherent functional bottlenecks of existing memory system in maintaining emotional intelligence across extended time spans, especially regarding the resolution of complex affective contradictions.
By introducing this novel dataset and benchmark, we establish a novel trajectory for advancing emotional intelligence within memory systems.
Future work will focus on advanced emotional processing and memory synchronization while simultaneously extending our framework to multimodal affective analysis to incorporate cross-modal behavioral cues for a more holistic emotional understanding.

% \section{Acknowledgments}

\bibliographystyle{ACM-Reference-Format}
\bibliography{main-base}

\newpage
\appendix

\begin{center}
    {\huge \textbf{Appendix}}
\end{center}

\section{Memory System Evaluation Workflow}
% 利用前面第五章构造的批量数据集，我们对记忆系统进行评估。我们设计了统一的评估流程，每个记忆系统都是分别从数据集中批量读取数据，然后执行内部的处理逻辑，产生结果输出。在将结果输出之后，我们利用统一的打分模板对各个记忆系统产生的结果进行打分。这样确保了对比的公平性。我们从数据读取到打分结果输出，设计了一套完整的评估流程，只需执行相应的评估脚本，即可得到评估结果。评估流程如\ref{fig:evaluate_workflow}所示。

\begin{figure*}[t]
    \centering
    \includegraphics[width=\linewidth]{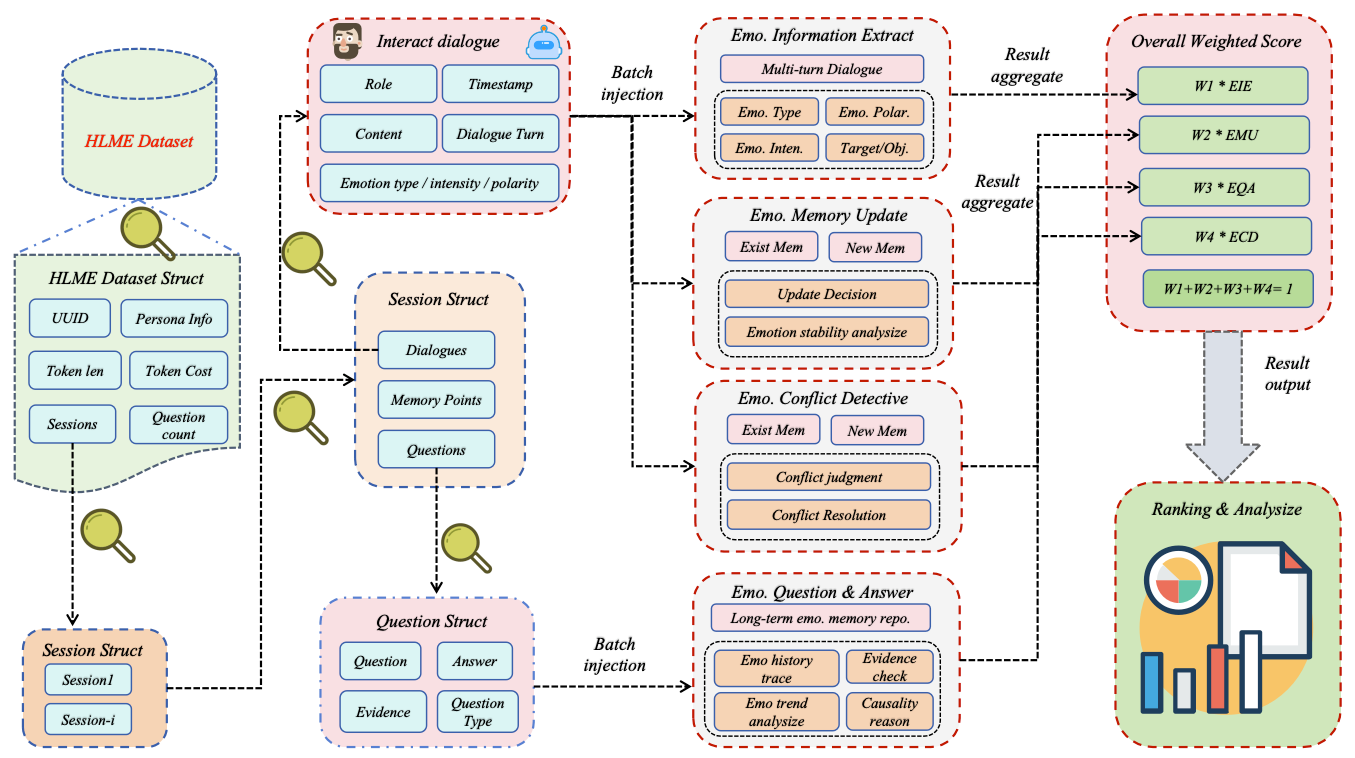}
    \caption{Workflow for batch emotion evaluation in memory systems.}
    \label{fig:evaluate_workflow}
\end{figure*}

Utilizing the multi-stage dataset constructed in Chapter 5, we conduct a systematic evaluation of various memory systems through a unified pipeline. Within this framework, each candidate system autonomously ingests records from the HLME dataset, executes its proprietary processing logic, and generates corresponding outputs. To ensure objectivity and cross-system comparability, we employ a standardized scoring template to assess all outputs. This comprehensive evaluation workflow, illustrated in Figure \ref{fig:evaluate_workflow}, encompasses the entire process from initial data loading to the final metric calculation. Quantitative results are efficiently derived by executing the designated evaluation scripts.

% 从图中可以看到，我们构建的HLME数据集包含六个元素，多个session组成了一个大的sessions集合，在session内部，对话作为重要的模块，被批量加载到记忆系统中，问题则在情感信息问答的时候被加载。评估打分系统都会将EIE, EMU, EQA, ECD四个任务的结果进行聚合。得到最终的加权输出。

As illustrated in the framework, the HLME dataset is organized into a hierarchical structure comprising six primary elements, where sessions are aggregated into a collective repository. Each session consists of multi-turn dialogues acting as core modules that are batch-loaded into the memory systems, with targeted queries injected during the Emotion Question \& Answer (EQA) phase. The evaluation framework quantifies performance across four critical dimensions: Emotion Information Extraction (EIE), Emotion Memory Update (EMU), EQA, and Emotion Conflict Detection (ECD). These dimensional scores are subsequently aggregated to produce a final weighted metric for comprehensive ranking and comparative analysis.

\section{Dataset Information}

\subsection{Individual dataset statistics}

We randomly selected a piece of data from the Medium and Large versions of the HLME dataset for preliminary analysis. The main statistical information is presented in Table \ref{tab:dataset_statistics}. Based on the structural information of individual data, we can calculate the size of the batch data set.

\begin{table}[htbp]
\centering
\setlength{\tabcolsep}{6pt}
\renewcommand{\arraystretch}{1.1}
\caption{A representative dataset instance and the statistical summary of its key components.}
\label{tab:dataset_statistics}
\resizebox{\columnwidth}{!}{
\begin{tabular}{lcc}
\toprule
\textbf{Metric} & \textbf{Medium} & \textbf{Large} \\
\midrule
Context Length (tokens)        & 170,304 & 1,000,229 \\
Session Num                    & 89 & 156 \\
Dialogue Turns                 & 2017 & 3169 \\
Memories Points Num                & 769 & 769 \\
Questions Num                & 234 & 234 \\

\midrule
\textit{Question Type Distribution} &  &  \\
Explicit Information Retrieval & 48 & 48 \\
Causal \& Logic Inference      & 30 & 30 \\
State Evolution Tracking       & 15 & 15 \\
Chronological Reasoning        & 7  & 7  \\
Memory Boundary                & 42 & 42 \\
Empathetic Generalization      & 45 & 45 \\
Conflict Detection \& Correction & 47 & 47 \\

\bottomrule
\end{tabular}
}
\end{table}

\subsection{Manual Annotation and Dataset Validation}
% 为了对HLME数据集的基本信息做一个简要分析，我们随机抽取了一条Medium版本数据和对应的Large版本数据进行了详细的统计。随机挑选的这条Medium版本单条数据包含2030轮对话，涵盖95个会话，对话文本总量为179155个Token。相比之下，Large版本将对话上下文扩展更大规模，在165个会话中包含3171轮对话，对话文本总量达到1000694个Token。尽管两个版本的上下文长度存在显著差异，但其评估核心保持了高度的一致性，每条典型数据均包含823个记忆点和260个精心设计的问题。

% 在问题类型的分布上，两个版本保持了统一的比例结构。具体分布为显性信息检索占比25.8%，冲突检测与纠正占比20.4%，共情泛化占比19.6%，记忆边界判定占比19.2%，状态演变追踪占比6.5%，因果与逻辑推理占比5.0%，以及时间顺序推理占比3.5%。这种设计确保了数据集中信息的密度和上下文的深度。

% 我们根据gpt-5-mini当前的计费标准计算了数据构造成本。该模型的每百万输入Token价格为0.25美元，每百万输出Token价格为2.00美元。在上面的这条随机抽取的数据中，Medium版本单条典型数据总计消耗7906081个输入Token和470240个输出Token，由此产生的单条数据构造成本为2.92美元。因此我们可以根据单条数据的构造成本，粗略计算整个数据集的经费开销。

% 在构建了HLME数据集之后,我们对数据集进行了检查。我们邀请了10位标注人员，他们分别具有教育、财务、体育、文学和医疗保健等领域的背景知识。我们从数据集中抽取了随机挑选了三分之一的数据条目，让标注人员分别对其medium和对应的large版本的数据集进行检查和标注。他们重点检查了每条对话包含的情绪标签信息、每个问题对应的答案和情感标签信息。通过这10位人员两两一组对数据集的交叉标注，我们确保了创建的HLME数据集的质量和情感的准确。为后续的大规模实验评估奠定了基础。

Following the construction of the HLME dataset, a multi-stage verification and quality control process was implemented to ensure data integrity and emotional precision. In the first stage, we recruited ten annotators with diverse expertise across education, finance, sports, literature, and healthcare. For this verification task, one-third of the total data entries were randomly selected, and the annotators reviewed both the \textit{Medium} and their corresponding \textit{Large} versions. This annotation focused on three critical dimensions encompassing the validity of emotional labels, the correctness of question answers, and the alignment of associated affective tags. To maximize inter-annotator agreement and mitigate individual bias, the ten participants were organized into five pairs for systematic cross-annotation and validation. 

In the second stage, five NLP experts specializing in sentiment analysis conducted an independent quality control audit on the processed dataset. Their independent validation demonstrated high human-machine agreement, exceeding 90\% on factual memory point annotation, 80\% on QA evidence retrieval, and 75\% on both fine-grained sentiment and event importance. These robust consistency rates collectively validate our incremental construction methodology and the high quality of the HLME dataset.

\section{Supplementary Experiments}

\subsection{EIE Emotional Intensity Analysis}

\begin{figure}[t]
    \centering
    \includegraphics[width=\columnwidth]{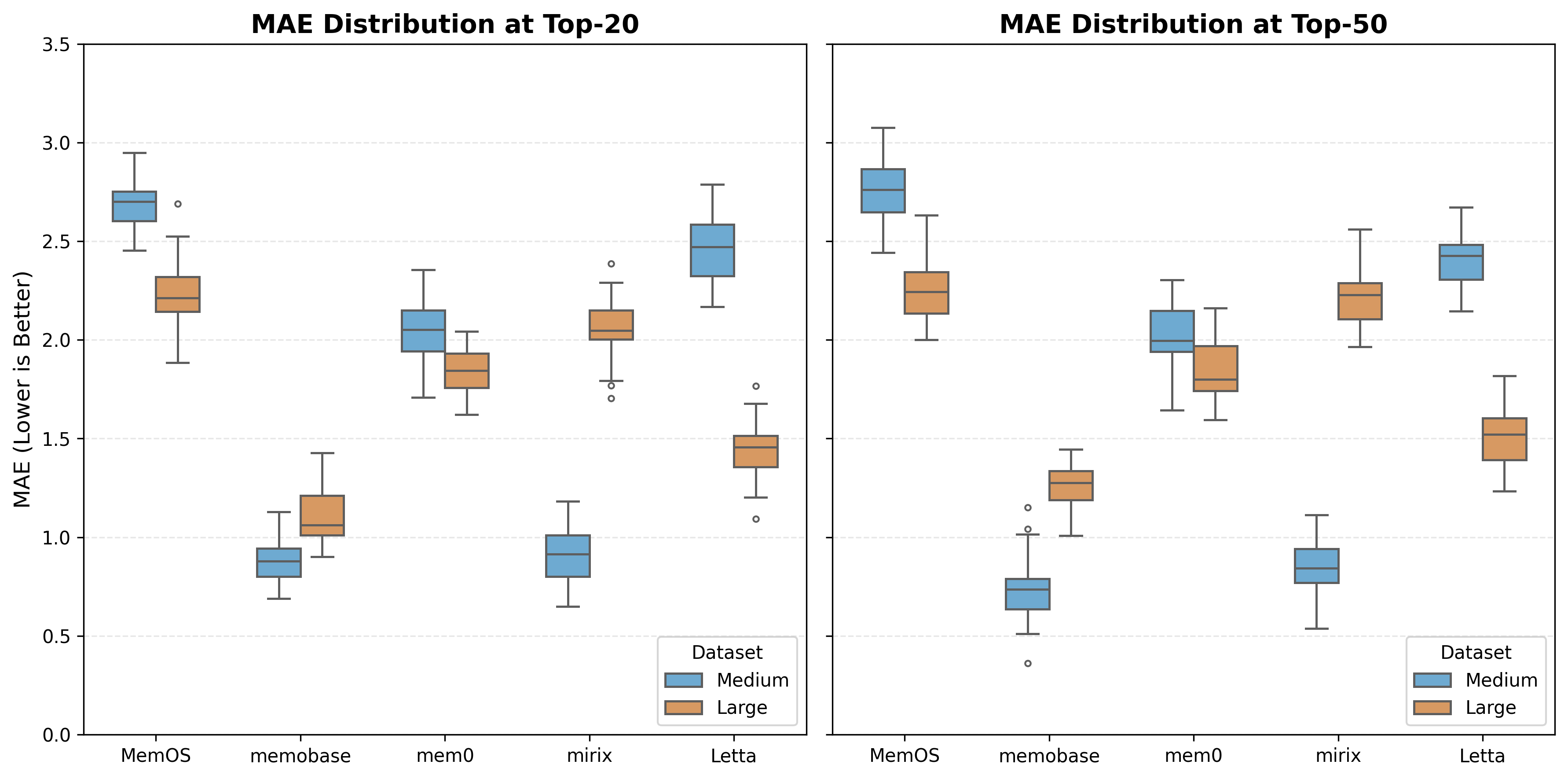}
    \caption{Distribution of Mean Absolute Error (MAE) for emotional intensity estimation in the EIE task across different retrieval depths. The boxplots compare the performance stability and error variance of memory systems on both Medium (blue) and Large (orange) datasets.}
    \label{fig:eie_mae_boxplot}
\end{figure}

To further evaluate the precision and stability of memory systems in estimating emotional intensity, we conduct a comparative analysis of the Mean Absolute Error (MAE) within the EIE task across different retrieval depths ($k=20$ and $k=50$), as illustrated in the boxplots of Figure \ref{fig:eie_mae_boxplot}.

A primary observation is the pronounced disparity in error distribution across the two dataset scales. Under both retrieval settings, the \textit{HLME-Medium} dataset (blue) generally exhibits more concentrated error profiles for most models, whereas the \textit{HLME-Large} dataset (orange) displays higher variance and a greater frequency of outliers. This trend suggests that as dialogue history expands, the escalating complexity of the emotional context compromises the models' ability to sustain stable intensity estimation.

Increasing the retrieval depth from $k=20$ to $k=50$ reveals divergent behaviors among the memory systems. For database-reliant architectures like \textit{MemoBase}, the MAE median on the \textit{Medium} dataset decreases, indicating that an expanded retrieval window successfully captures more relevant emotional cues to calibrate intensity perception. Conversely, for reasoning-intensive systems like \textit{MemOS}, the error floor remains elevated at an MAE of approximately 2.7 and exhibits a slight upward shift in the Top-50 setting, validating the distractor effect where extra retrieved items introduce noise that impedes fine-grained reasoning.

Furthermore, \textit{Letta} demonstrates remarkable resilience across dataset scales and retrieval depths, maintaining a stable MAE on the \textit{HLME-Large} dataset to highlight its structural robustness in managing extensive context windows. In contrast, the efficacy of \textit{Mirix} is highly contingent on scale; its MAE undergoes a near-fold increase when transitioning to the \textit{Large} setting. This gap suggests that while \textit{Mirix} excels in constrained contexts, its pipeline struggles with error propagation as emotional complexity intensifies, whereas \textit{Letta} provides a more reliable anchor for long-term affective tracking.

\subsection{EMU Emotional Intensity Analysis}

\begin{figure}[htbp]
    \centering
    \includegraphics[width=\columnwidth]{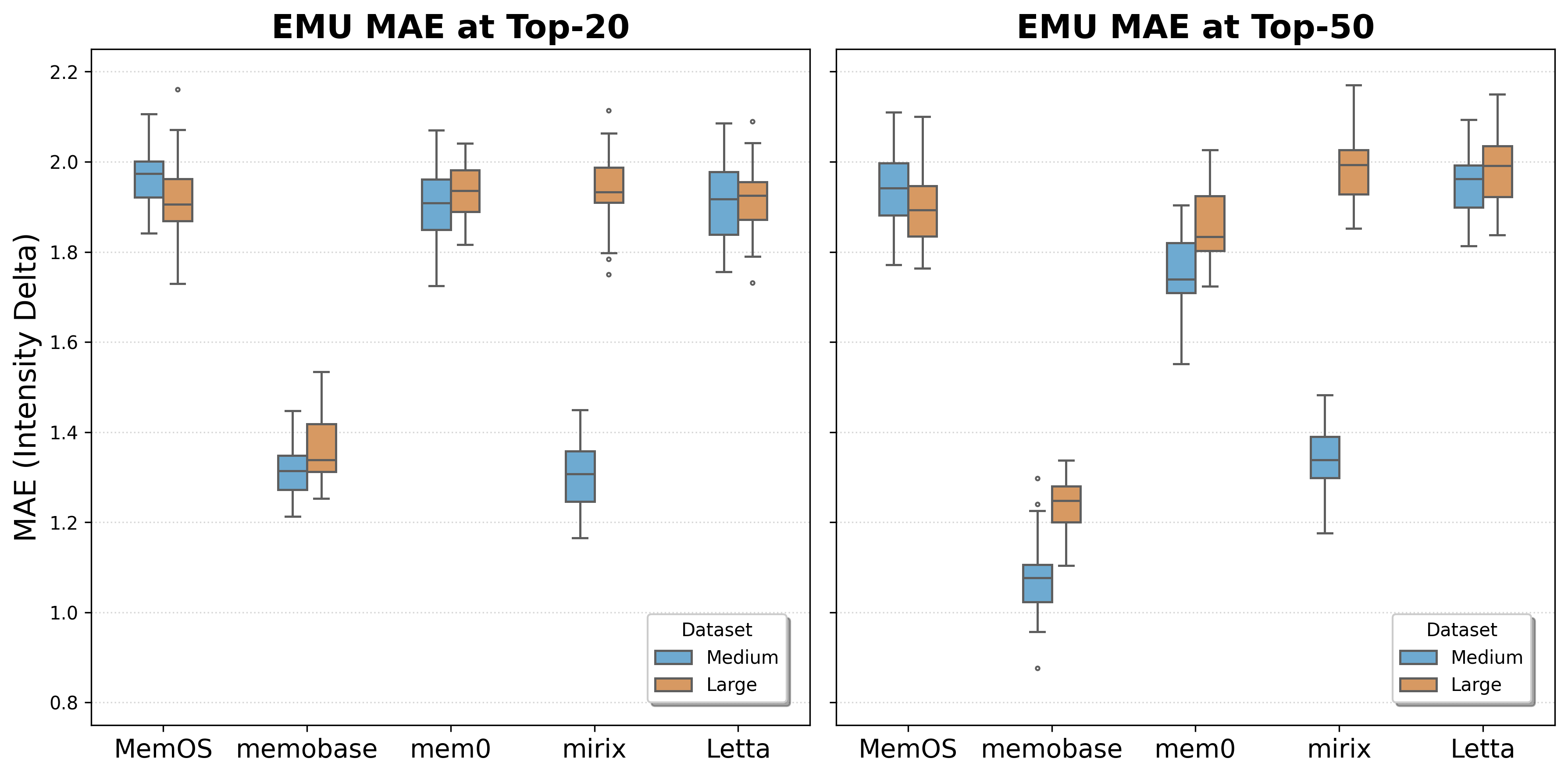}
    \caption{$MAE_\Delta$ metrics in different top-k for EMU task.}
    \label{fig:emu_mae}
\end{figure}

The $MAE_\Delta$ metric for the EMU task assesses the system ability to synchronize internal memory with evolving emotional states as illustrated in Figure \ref{fig:emu_mae}. MemoBase and Mirix demonstrate superior precision on the Medium dataset with $MAE_\Delta$ values of $1.3289$ and $1.3026$ respectively at a retrieval depth of twenty. MemoBase exhibits unique scaling advantages as its $MAE_\Delta$ improves to $1.0921$ at a depth of fifty on the Medium scale and reaches $1.2368$ on the Large scale. This sustained performance is driven by the use of local database calls which offer higher stability than the remote API communication used by other memory systems.

The sharp increase in $MAE_\Delta$ for Mirix on the Large dataset suggests that extensive context noise impairs its ability to track intensity fluctuations. The observed trends indicate that while increasing retrieval depth benefits systems with robust filtering mechanisms, it may lead to significant error accumulation in memory systems that are highly sensitive to long range noise.

\subsection{Statistical Correlation Analysis of Evaluation Metrics}

\begin{figure}[htbp]
    \centering
    \includegraphics[width=\columnwidth]{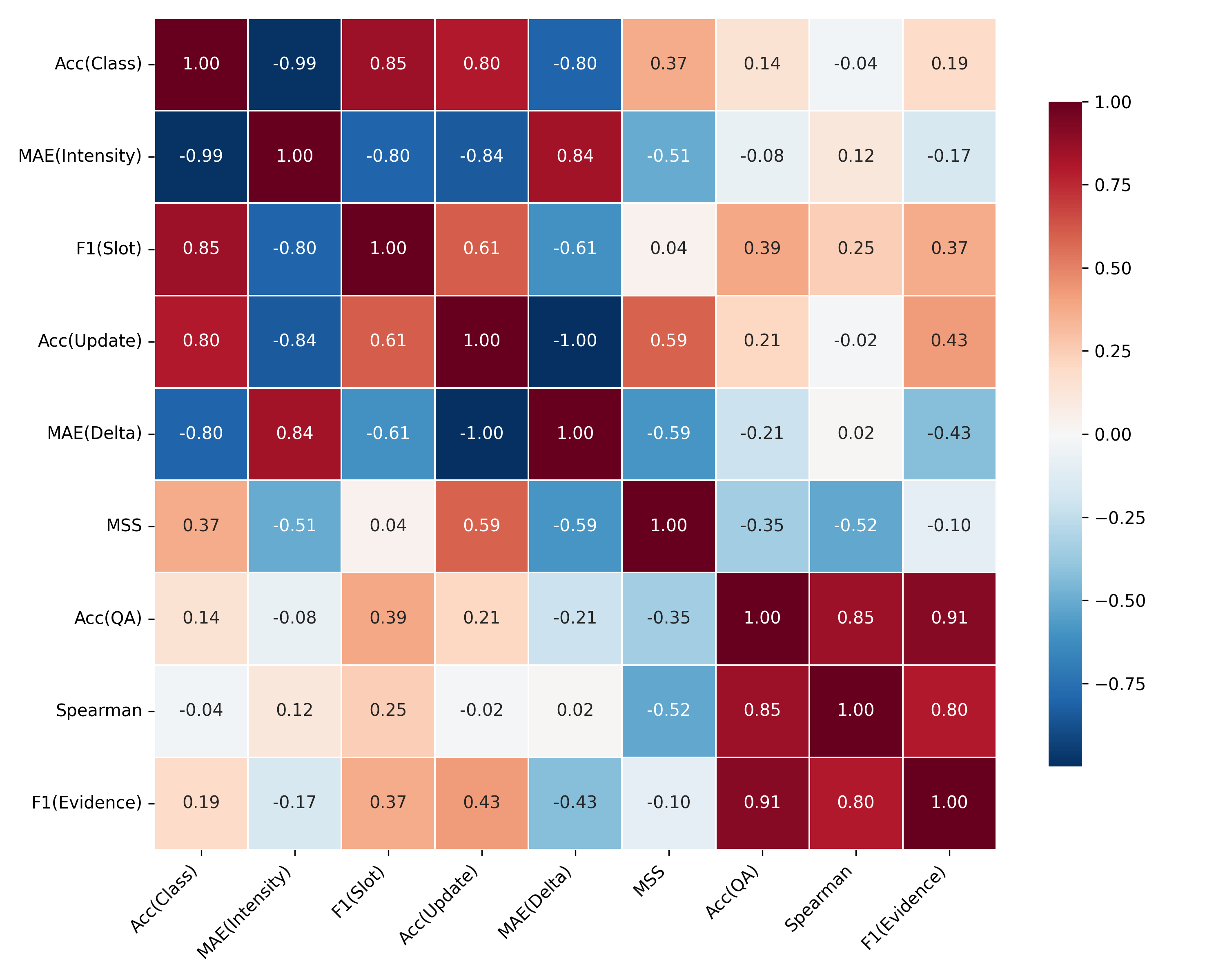}
    \caption{Statistical correlation analysis of metrics.}
    \label{fig:result_related}
\end{figure}

The correlation analysis illustrated in Figure \ref{fig:result_related} reveals critical metrics dependencies within the MemEmo benchmark. A strong categorical-intensity coupling is observed, where classification accuracy and mean absolute error ($MAE$) are inversely correlated; this indicates that precise emotional quantification is strictly contingent upon correct category identification. This is complemented by a significant extraction-update dependency, where the positive correlation between EIE slot F1 scores and EMU update accuracy confirms that robust information extraction serves as a functional prerequisite for successful memory synchronization.

Furthermore, the weak correlation between EQA accuracy and intensity delta error ($MAE_\Delta$) suggests a reasoning-memory decoupling, where downstream tasks rely more on the LLM’s inherent semantic reasoning than on stored numerical precision. Finally, the strong positive correlation between evidence F1 and Spearman's $\rho$ validates that the ability to retrieve supporting evidence is intrinsically linked to the logical coherence of emotional trends.

\subsection{Multi-task Performance under Top-10 Contextual Retrieval}
\begin{figure}[!htbp]
    \centering
    \includegraphics[width=\columnwidth]{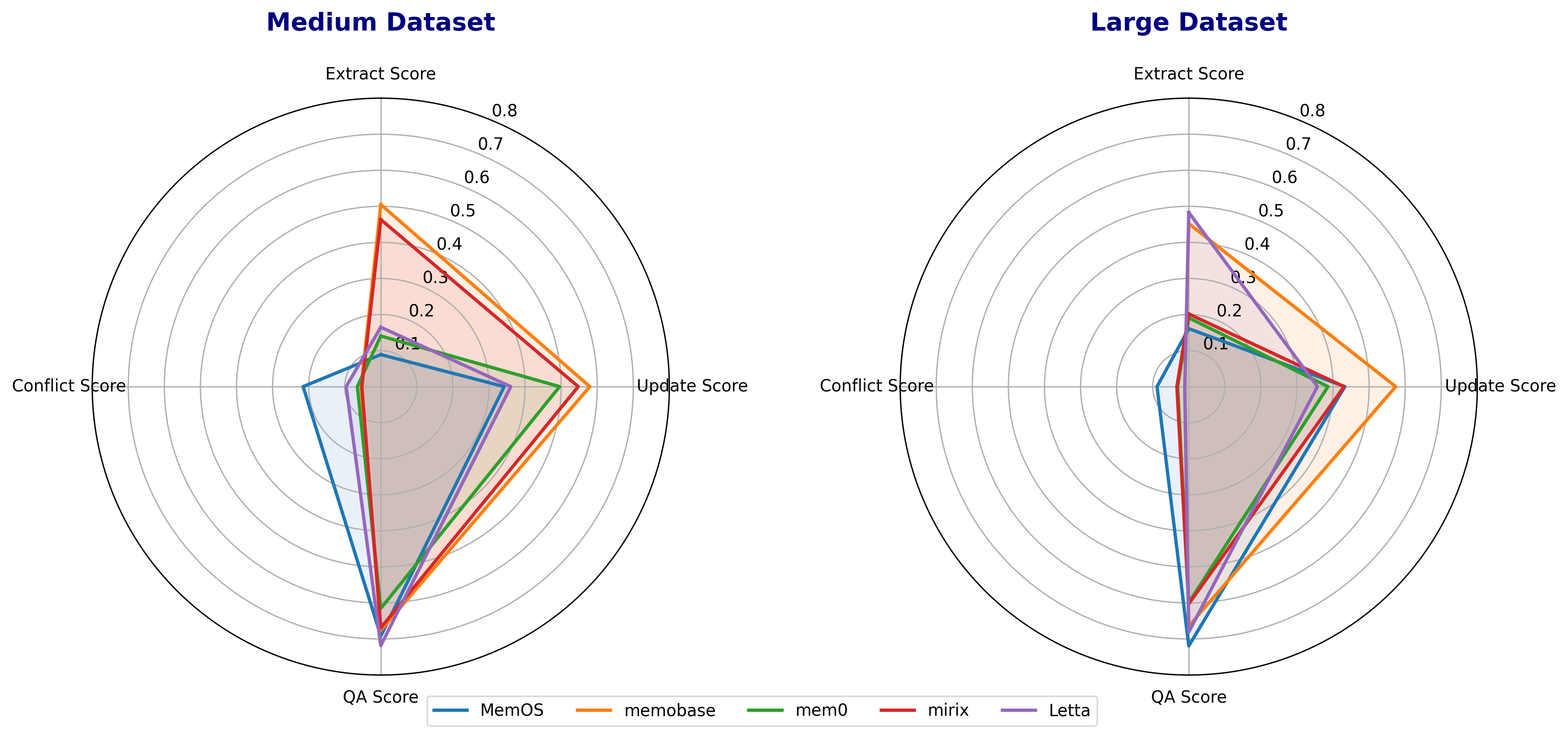}
    \caption{Comparative analysis of memory systems across EIE, EMU, EQA, and ECD tasks on the Medium (left) and Large (right) datasets under the Top-10 retrieval setting.}
    \label{fig:radar_comparison}
\end{figure}

To further characterize the functional profiles of the evaluated memory systems, we visualize their task-specific capabilities using radar charts across two dataset scales, as illustrated in Figure \ref{fig:radar_comparison}.

On the HLME-Medium dataset, the candidate systems display highly specialized performance profiles. \textit{MemoBase} and \textit{Mirix} exhibit superior proficiency in the Emotion Information Extraction (EIE) and Emotion Memory Update (EMU) dimensions. Conversely, \textit{MemOS} demonstrates a unique structural advantage in Emotion Conflict Detection (ECD), maintaining significantly broader coverage in the corresponding radar chart quadrant compared to its counterparts. \textit{Letta} presents a more balanced profile, with a notable strength in the Emotion Question \& Answer (EQA) task.

Turning to the HLME-Large dataset, a general contraction in the radar surface area is observed for several models, indicating a performance degradation as the dialogue history and emotional complexity increase. Notably, \textit{Letta} shows remarkable robustness in EIE even at scale, while \textit{MemoBase} maintains its dominance in memory updating. However, the ECD task remains a significant bottleneck for almost all systems in the large-scale setting, with the corresponding vertices collapsing toward the origin. This visualization confirms that while current systems excel at static extraction and retrieval, managing dynamic emotional conflicts and maintaining long-term stability remains a non-trivial challenge.

Radar profile analysis reveals a clear trade-off between specialized proficiency and overall robustness across memory systems. While MemoBase and Mirix lead in emotional extraction and updating on the medium scale, MemOS distinguishes itself through superior conflict detection capabilities. However, the transition to the HLME-Large dataset exposes a general performance contraction, particularly in Emotion Conflict Detection (ECD), which remains a critical bottleneck for all evaluated systems. These results reveal that while current memory systems are effective for static information retrieval, maintaining emotional consistency and resolving complex affective contradictions in large-scale long-term memory remains a formidable challenge.

\section{Emotional Configuration and Design Logic}

\subsection{Emotion Taxonomy}

We adopt the 49 emotion categories defined in the Emotion Annotation and Representation Language (EARL) from the HUMAINE framework. To enable fine-grained modeling, we extend this taxonomy into a structured configuration schema where each emotion $e$ is represented as a tuple:
\[
e = (c, t, \mathcal{I}, \mathcal{T})
\]
Where $c$ denotes the emotion category, $t \in \{\text{positive}, \text{negative}\}$ represents the polarity, $\mathcal{I}$ is a set of discrete intensity tags, and $\mathcal{T}$ defines the corresponding threshold mappings.

The intensity of each emotion is modeled through a three-level hierarchy (\textit{Low}, \textit{Medium}, \textit{High}). Each level is associated with a normalized tag for surface realization and a natural language description for interpretability. Formally, the intensity mapping is defined as:
\[
\mathcal{T} = \{(\theta_1, \text{High}), (\theta_2, \text{Medium}), (\theta_3, \text{Low})\}
\]
where $\theta_1 > \theta_2 > \theta_3$ are empirically defined thresholds. This design bridges discrete semantic anchors with continuous modeling of emotional magnitude.

The 49 EARL categories are organized into the following semantically coherent groups, serving as the basis for our affective modeling:

% \begin{itemize}
    \textbf{Negative States:} 
    \begin{itemize}
        \item \textit{Forceful:} anger, annoyance, contempt, disgust, irritation.
        \item \textit{Lack-of-control:} anxiety, embarrassment, fear, helplessness, powerlessness, worry.
        \item \textit{Cognitive:} doubt, envy, frustration, guilt, shame.
        \item \textit{Passive:} boredom, despair, disappointment, hurt, sadness, agitation, stress, shock, tension.
    \end{itemize}

    \textbf{Positive States:} 
    \begin{itemize}
        \item \textit{Lively:} amusement, delight, elation, excitement, happiness, joy, pleasure, caring, affection, empathy, friendliness, love.
        \item \textit{Cognitive:} pride, courage, hope, humility, satisfaction, trust.
        \item \textit{Quiet:} calmness, contentment, relaxation, relief, serenity.
    \end{itemize}

    \textbf{Reactive States:} interest, politeness, surprise.
% \end{itemize}

\textbf{Dynamic Calibration of Reactive States.} 
Within our framework, reactive states including \textit{interest}, \textit{politeness}, and \textit{surprise} are assigned a default positive polarity during the initial configuration phase. This preset serves as a functional bias reflecting the prosocial nature of standard human-computer interaction. However, during the subsequent dialogue generation process, the Large Language Model (LLM) performs a secondary calibration to synchronize the relationships between the event descriptions, memory point, and affective states. If the semantic context of an event manifests a pessimistic or adverse orientation, the system automatically rectifies the bound reactive emotion to its corresponding negative polarity. This multi-stage refinement ensures that the association between conversational content and emotional information remains contextually accurate and logically consistent.

\subsection{Structured Emotional Configuration}

To operationalize the HUMAINE EARL emotion taxonomy in a machine-interpretable format, we design a structured configuration schema, as exemplified in Listing \ref{lst:emotional_config}. Each emotion is defined as a structured entry comprising a unique identifier, a canonical name, and its corresponding polarity.

Specifically, each category incorporates a three-level intensity hierarchy comprising Low, Medium, and High states. Each level is associated with a normalized lexical tag and a natural language description to capture subtle variations in emotional expression. For example, the emotion \textit{anger} is refined into \textit{irritated}, \textit{angry}, and \textit{rage} to reflect increasing stages of emotional strength. In our emotion configuration, the canonical emotion category terms can be substituted with synonyms representing moderate-intensity affective labels.

Furthermore, the configuration defines explicit threshold mappings that connect continuous numerical scores to discrete intensity levels. This mechanism allows the system to integrate scalar emotion predictions while preserving the interpretability of discrete labels. 

As illustrated in the JSON snippet, these structured fields facilitate a unified representation that bridges discrete categories with continuous intensity modeling, thereby supporting fine-grained affective analysis across multiple evaluation dimensions.

\begin{lstlisting}[language=json, caption={Example of structured emotional configuration}, label={lst:emotional_config}]
"Emotions_HUMAINE_EARL": [
{
    "id": 1,
    "name_en": "anger",
    "emotion_type": "negative",
    "intensity_tags": [
    { 
        "level": "High", "tag": "rage", 
        "description": "Intense and explosive anger potentially out of control" 
    },
    { 
        "level": "Medium", "tag": "angry", 
        "description": "Clearly angry and dissatisfied" 
    },
    { 
        "level": "Low", "tag": "irritated", 
        "description": "Slightly annoyed" 
    }
    ],
    "intensity_thresholds": [[0.7, "High"],[0.3, "Medium"], [0.0, "Low"]]
},
{
    "id": 2,
    "name_en": "annoyance",
    "emotion_type": "negative",
    "intensity_tags": [
    { 
        "level": "High", "tag": "fed_up", 
        "description": "Strong aversion feeling" 
    },
    { 
        "level": "Medium", "tag": "annoyed", 
        "description": "Persistent irritability" 
    },
    { 
        "level": "Low", "tag": "bothered", 
        "description": "Mild discomfort" 
    }
    ],
    "intensity_thresholds": [[0.7, "High"],[0.3, "Medium"], [0.0, "Low"]]
},...
]
\end{lstlisting}

\subsection{Design Logic of Emotional Transition Rules}

To evaluate the affective modeling module of system, this study designs a set of multi-dimensional Emotional Transition Rules as detailed in Listing \ref{lst:emotional_trans}. These rules are specifically developed to simulate the evolution of emotional trajectories across core life domains comprising career shifts, health conditions, interpersonal relationships, annual plan progress, and preference adjustments.

% \textbf{1. Event-Driven Emotion Induction Mechanism.} 
The proposed rule system adopts event-based emotion appraisal theories to model affective responses to environmental stimuli. Under the \textit{career\_change} category, a promotion is mapped to feelings of pride and joy, which aligns with the sense of self-actualization following professional achievement. Conversely, the rules for \textit{pressure} and \textit{failure} capture the cascading reactions of negative emotions such as worry and shame. Within the \textit{relationship\_change} module, the rules distinguish between positive affect arising from \textit{bonding} and the anger or despair resulting from \textit{conflict} or a \textit{break\_up}. This differentiation reflects the decisive role of social belonging in maintaining psychological homeostasis.

% \textbf{2. Goal-Oriented and Cognitive Appraisal-Based Dynamic Adjustment.} 
In the design of \textit{annual\_plan\_progress} and \textit{preference\_change}, the rules incorporate cognitive appraisal theory to handle internal goal states. Falling \textit{behind} on a plan induces not only stress but also a deeper sense of powerlessness. Within the logic of preference deletion designated as \textit{Delete}, adjustments achieved through \textit{self\_clarification} are imbued with the emotional attributes of relief and acceptance rather than being treated as mere passive abandonment. This design simulates the human psychological process of mitigating cognitive dissonance and achieving emotional restoration by realigning goals during self-management.

% \textbf{3. Coexistence and Evolution of Multiple Emotional States.} 
Each triggering event is configured with three emotional labels to reflect that emotions are complex affective clusters rather than isolated states. For example, an improvement in health denoted as \textit{health\_change: improve} is simultaneously accompanied by relief, contentment, and calmness. This multi-dimensional mapping enhances the memory system's ability to process affective granularity during subsequent reasoning tasks and ensures a more nuanced representation of the human emotional experience.

\begin{lstlisting}[language=json, caption={Emotional transition rules}, label={lst:emotional_trans}]
"Emotion_Transition_Rules" :{
    "career_change": {
        "promotion": ["pride", "joy", "hope"],
        "improve": ["satisfaction", "hope", "trust"],
        "pressure": ["stress", "tension", "worry"],
        "failure": ["frustration", "shame", "doubt"]
    },
    "health_change": {
        "improve": ["relief", "contentment", "calmness"],
        "illness": ["fear", "powerlessness", "helplessness"],
        "chronic": ["sadness", "powerlessness", "worry"]
    },
    "relationship_change": {
        "bonding": ["love", "friendliness", "happiness"],
        "conflict": ["anger", "hurt", "sadness"],
        "break_up": ["despair", "shame", "sadness"],
        "support_gain": ["empathy", "trust", "serenity"]
    },
    "annual_plan_progress": {
        "on_track": ["excitement", "satisfaction", "contentment"],
        "behind": ["shame", "stress", "powerlessness"],
        "failed": ["disappointment", "despair", "guilt"]
    },
    "preference_change": {
        "Delete": {
            "self_clarification": ["relief","acceptance","contentment"],
            "goal_refocus": ["determination", "confidence"],
            "constraint_adjustment": ["resignation", "calmness"]
        },
        "Modify": {
            "optimization": ["satisfaction", "confidence", "hope"],
            "upgrade": ["anticipation", "excitement", "hope"],
            "fine_tuning": ["contentment", "calmness"]
        }
    }
}
\end{lstlisting}

\section{Evaluation System Prompt Design}
% 为了对前面提到的5个记忆系统进行评估，我们设计了统一的提示模板来进行情感信息抽取，情感记忆更新，情感问答和情感冲突检测等四个维度的评估。我们使用LLM-as-Judge来判断当前这些主流的记忆系统在这四个维度上的表现。我们设计的四个提示模板如下面的四个小节所示。

To evaluate the five memory systems mentioned previously, we designed a unified prompt template to conduct assessments across four dimensions comprising emotional information extraction, emotional memory updating, emotional question answering, and emotional conflict detection. We leverage the \textit{LLM-as-Judge} paradigm to evaluate the performance of these mainstream memory systems across the specified dimensions. The design of the corresponding prompt templates is detailed in the following four subsections.

\subsection{Emotion Information Extraction Prompt}
% 记忆系统在执行完情感信息抽取任务之后会产生相应的输出，不同的记忆系统产生的输出各不相同。为了评估各个记忆系统在情感信息抽取方面的能力表现，我们设计了EIE 评估模板来统一评估各个记忆系统的得分表现，评估模板设计如下所示:

Different memory systems generate heterogeneous outputs after the emotional information extraction process. To quantify their capabilities, we developed a unified EIE evaluation template.

\begin{tcolorbox}[
    colback=blue!5,
    colframe=blue!50!black,
    title=EIE Evaluation Prompt,
    fonttitle=\bfseries,
    breakable
]

\textbf{Role:} \\
You are an \textbf{Emotion Information Extraction (EIE) Evaluator}. Your task is to compare the \textbf{System Extracted Emotion Memory} against the \textbf{Ground Truth Emotion Memory}.

\vspace{0.5em}
\textbf{Input Data:}
\begin{itemize}
    \item Dialogue Context: \texttt{\{dialogue\}}
    \item Ground Truth Memory: \texttt{\{gold\_memory\}}
    \item System Extracted Memory: \texttt{\{sys\_output\}}
\end{itemize}

\vspace{0.5em}
\textbf{Evaluation Rules:}
\begin{itemize}
    \item \textbf{Intensity Mapping:}
    \begin{itemize}
        \item High / Very / Strong / Level 3 $\rightarrow$ 3
        \item Medium / Normal / Moderate / Level 2 $\rightarrow$ 2
        \item Low / Slight / Mild / Level 1 $\rightarrow$ 1
    \end{itemize}
    \item \textbf{MAE Penalty:} 
    If the emotion category is correct but intensity differs significantly, ensure that \texttt{intensity\_diff} reflects the full numerical gap.
\end{itemize}

\vspace{0.5em}
\textbf{Evaluation Tasks:}
\begin{itemize}
    \item Task 1: Emotion Type Classification (Boolean)
    \item Task 2: Intensity Estimation (Numerical Difference)
    \item Task 3: Slot Filling F1 (Float)
\end{itemize}

\vspace{0.5em}
\textbf{Evaluation Logic:}
\begin{enumerate}
    \item \textbf{type\_match:} \\
    Whether the extracted emotion category matches the Ground Truth. Semantic equivalence is allowed.

    \item \textbf{intensity\_diff:} \\
    Map intensity to numerical values (Low=1, Medium=2, High=3), and compute:
    \texttt{intensity\_diff = |System - Gold|}. \\
    If no emotion is extracted, set \texttt{intensity\_diff = 3.0}.

    \item \textbf{slot\_f1:} \\
    A slot is defined as \texttt{<Emotion Category, Polarity, Target/Object>}. \\
    Compute Precision, Recall, and F1 score based on extracted and ground truth slots. Semantic equivalence is allowed.
\end{enumerate}

\vspace{0.5em}
\textbf{Output Format (JSON ONLY):}

\begin{verbatim}
{
    "type_match": <Boolean>,
    "intensity_diff": <Float>,
    "slot_f1": <Float>,
    "reasoning": "<String>"
}
\end{verbatim}

\end{tcolorbox}

\subsection{Emotion Memory Update Prompt}
% 记忆系统在添加信息的时候首先要判断该条信息是否已经存在在系统中，对于已经存在的信息，还要决定是否进行更新操作。EMU评估模板为各个记忆系统提供了统一的评估标准。

When incorporating new data, memory systems are required to detect information redundancy and manage the subsequent update logic for existing records. The EMU evaluation template establishes a unified metric for benchmarking the updating efficiency and accuracy of various memory systems.

\begin{tcolorbox}[
    colback=blue!5,
    colframe=blue!50!black,
    title=EMU Evaluation Prompt,
    fonttitle=\bfseries,
    breakable
]

\textbf{Role:} \\
You are an \textbf{Emotion Memory Update (EMU) Evaluator}. Your task is to assess whether the system correctly updates memory by comparing system states before and after the update.

\vspace{0.5em}
\textbf{Input Data:}
\begin{itemize}
    \item System Memory BEFORE (State t): \texttt{\{original\_memory\}}
    \item Gold New State (State t+1): \texttt{\{gold\_new\_state\}}
    \item System Memory AFTER (State t+1): \texttt{\{sys\_output\}}
\end{itemize}

\vspace{0.5em}
\textbf{Evaluation Logic:}
\begin{enumerate}
    \item \textbf{decision\_correct:} \\
    Compare System AFTER with System BEFORE. Determine whether the system correctly updates the target emotion to match the Gold New State. \\
    Return False if the system memory remains unchanged or the target information is missing.

    \item \textbf{delta\_diff:} \\
    Map intensity values as Low=1, Medium=2, High=3, None=0. \\
    Compute: \texttt{|Gold\_t+1 - System\_t+1|}. \\
    If \texttt{decision\_correct = False}, assign a penalty: \texttt{delta\_diff = 2.0}.

    \item \textbf{is\_stable\_correct:} \\
    Determine whether unrelated memory entries from System BEFORE are preserved in System AFTER.
\end{enumerate}

\vspace{0.5em}
\textbf{Output Format (JSON ONLY):}

\begin{verbatim}
{
    "decision_correct": <Boolean>,
    "delta_diff": <Float>,
    "is_stable_correct": <Boolean>,
    "reasoning": "<Step-by-step logic>"
}
\end{verbatim}

\end{tcolorbox}

\subsection{Emotion Question \& Answer Prompt}
% 数据集被加载到记忆系统之后，开始进行问答推理。对于同一个问题，不同的记忆系统给出了不同的答案。为了对各个记忆系统的问题回答进行统一评判。我们设计了EQA评估提示模板。

To reconcile the variability in the responses of the memory system, the EQA evaluation template provides a standardized protocol for consistent evaluation of reasoning performance.

\begin{tcolorbox}[
    colback=blue!5,
    colframe=blue!50!black,
    title=EQA Evaluation Prompt,
    fonttitle=\bfseries,
    breakable
]

\textbf{Role:} \\
You are an \textbf{Emotion Q\&A Evaluator}. Your task is to assess whether the system correctly answers questions about emotional memory, including emotional states, temporal changes, and underlying causes.

\vspace{0.5em}
\textbf{Input Data:}
\begin{itemize}
    \item Question: \texttt{\{question\}}
    \item Reference Answer (Gold): \texttt{\{gold\_answer\}}
    \item System Response: \texttt{\{sys\_response\}}
\end{itemize}

\vspace{0.5em}
\textbf{Evaluation Tasks:}
\begin{itemize}
    \item Task 1: Emotion QA Accuracy
    \item Task 2: Emotion Intensity Trend Extraction
    \item Task 3: Memory Evidence Grounding F1
\end{itemize}

\vspace{0.5em}
\textbf{Evaluation Logic:}
\begin{enumerate}
    \item \textbf{is\_correct:} \\
    Compare the system response with the gold answer at the factual level. \\
    True if the answer is factually correct or semantically equivalent. \\
    False if the system provides incorrect emotional facts, wrong temporal judgments, or fails to answer meaningfully. \\
    This metric evaluates correctness only and does not penalize missing explanations.

    \item \textbf{pred\_intensity\_trend:} \\
    Identify emotional intensity changes over time from the system response. \\
    Map intensity using: Low=1, Medium=2, High=3, None=0. \\
    Output a list representing the temporal trajectory. \\
    Example: \texttt{[3, 0]} for a transition from strong emotion to none.

    \item \textbf{evidence\_f1:} \\
    Identify supporting emotional facts used by the system and compare them with the gold answer. \\
    Compute Precision, Recall, and F1. \\
    Apply a penalty if hallucinated or irrelevant evidence is used.
\end{enumerate}

\vspace{0.5em}
\textbf{Output Format (JSON ONLY):}

\begin{verbatim}
{
    "is_correct": <Boolean>,
    "pred_intensity_trend": <List of Integers>,
    "evidence_f1": <Float>,
    "reasoning": "<Explanation>"
}
\end{verbatim}

\end{tcolorbox}

\subsection{Emotion Conflict Detection Prompt}
% 情感冲突的判定比情感记忆更新提高了一个难度等级。需要在更新的内容中判定是否存在冲突。ECD评估模板为各个记忆系统的冲突判定提供了一个统一的判定标准。

ECD increases task complexity by requiring systems to identify logical incongruities within updated content. ECD templates provide a uniform standard for conflict detection across different systems.

\begin{tcolorbox}[
    colback=blue!5,
    colframe=blue!50!black,
    title=ECD Evaluation Prompt,
    fonttitle=\bfseries,
    breakable
]

\textbf{Role:} \\
You are an \textbf{Emotion Conflict Detection (ECD) Evaluator}. Your task is to determine how the memory system handles conflicting emotional information by comparing memory states before and after a new dialogue.

\vspace{0.5em}
\textbf{Input Data:}
\begin{itemize}
    \item New Dialogue (Trigger): \texttt{\{dialogue\}}
    \item Gold Conflict Strategy Code: \texttt{\{gold\_conflict\}} \\
    (0 = No Conflict, 1 = Overwrite, 2 = Label, 3 = Coexist)
    \item Memory BEFORE Dialogue: \texttt{\{mems\_before\}}
    \item Memory AFTER Dialogue: \texttt{\{mems\_after\}}
\end{itemize}

\vspace{0.5em}
\textbf{Evaluation Logic:}
\begin{enumerate}
    \item \textbf{detected\_by\_system:} \\
    Determine whether the system actively handles the conflict. \\
    True if the system overwrites previous memory or explicitly labels the conflict. \\
    False if the system blindly appends conflicting information or fails to record the new emotion.

    \item \textbf{strategy\_code\_executed:} \\
    Identify the strategy used:
    \begin{itemize}
        \item 1 (Overwrite): Old emotion is fully replaced by the new one
        \item 2 (Label): Both exist with explicit indication of conflict or change
        \item 3 (Coexist): Both exist without any conflict resolution logic
    \end{itemize}
    If no effective action is taken, default to 3.

    \item \textbf{strategy\_match:} \\
    True if the executed strategy matches the gold conflict strategy; otherwise False.
\end{enumerate}

\vspace{0.5em}
\textbf{Output Format (JSON ONLY):}

\begin{verbatim}
{
    "detected_by_system": <Boolean>,
    "strategy_code_executed": <Integer>,
    "strategy_match": <Boolean>,
    "reasoning": "<Step-by-step analysis>"
}
\end{verbatim}

\end{tcolorbox}

\section{Key Templates for Dataset Construction}

% 定义 Prompt 方框样式
\newtcolorbox{promptbox}[1]{
    colback=gray!5!white,
    colframe=gray!75!black,
    fonttitle=\bfseries\sffamily,
    coltitle=white,
    title=#1,
    enhanced,
    attach boxed title to top left={yshift=-2mm, xshift=2mm},
    boxed title style={colframe=green!75!black, colback=green!75!black},
    sharp corners,
    boxrule=0.5pt,
    breakable,
    fontupper=\small\sffamily
}

% 在第四章中，我们详细介绍了数据集的构造方法。在构建数据集的过程中，我们在关键的阶段都设计了prompt模板，通过LLM的介入来检查和优化我们的构建流程。本章后续的小节将简要介绍几个构建过程中的关键prompt设计。完整的prompt和代码在我们提供的代码压缩包中展示。通过执行相关的配置之后可以批量化的数据集构造和实验评估。

In Chapter 4, we provide a detailed description of the methodology employed for dataset construction. Throughout the construction process, specialized prompt templates are designed at critical stages to facilitate the inspection and optimization of our workflow through Large Language Model (LLM) intervention. Subsequent sections of this chapter briefly introduce several key prompt designs utilized during development. The complete prompts and implementation details are presented in the code package. Following the execution of relevant configurations, this framework enables large-scale automated dataset construction and experimental evaluation.

% 本章将生成对话准备模板、冲突情感生成模板，对话生成模板和问题生成模板进行简单的介绍。这些关键的模板是我们构建情感记忆数据集的关键模板，在我们生成数据集的过程中发挥了重要作用。

% We present a brief introduction to the key templates used for dialogue preparation, conflicting emotion synthesis, and the generation of conversations and queries. These templates constitute the essential framework for the HLME dataset.

\subsection{Generate Conversation Preparation}
% 为了将离散的生活事件转化为结构化的对话上下文，我们设计了如下方框所示的对话信息生成模板。该模板作为一个分析引擎，能够综合用户画像、历史事件日志和当前事件刺激，生成连贯的对话目标和摘要。该设计的一个显著特征是能够提取具有量化重要性评分（从日常琐事到核心决策）的记忆点。该模板确保生成的对话数据不仅反映了自然的人际互动，而且可以作为评估记忆系统在长期用户画像维护和情感细微差别校准能力方面的精准基准。

To transform discrete life events into structured conversational contexts, we developed a comprehensive dialogue information generation template, as detailed in the following prompt box. This template functions as an analysis engine that synthesizes character profiles, historical logs, and current event stimuli to generate coherent dialogue goals and summaries. A distinguishing feature of this design is its ability to distill salient memory points and generate concise dialogue summaries ranging from daily trivialities to pivotal life decisions. The template ensures that the synthesized conversational data not only reflects natural human interaction but also serves as a robust baseline for evaluating a memory system's capacity for emotional nuance calibration.

\begin{promptbox}{Conversation Generation Preparation Template}
    \textbf{Role Description} \\
    You are a top-tier dialogue generation and analysis engine. Your core task is to generate high-quality dialogue summaries based on provided background and event information, and extract structured memory data to maintain an accurate user profile.

    \vspace{2mm}
    \textbf{Task Description} \\
    The input includes character profiles, historical events, stored memories, current events, and motivation types. You are required to generate:
    \begin{itemize}
        \item Temporal boundaries (start and end time points) and dialogue goals.
        \item A comprehensive dialogue summary rich in emotional and contextual detail.
        \item A secondary memory point list including index, content, type, and update status.
    \end{itemize}

    \vspace{2mm}
    \textbf{Memory Type \& Importance Evaluation} \\
    Memories are categorized into \textit{Persona}, \textit{Event}, and \textit{Relationship}. Importance is quantified as:
    \begin{itemize}
        \item \textbf{0.25}: Daily trivialities or temporary thoughts.
        \item \textbf{0.50}: Significant information with limited long-term impact.
        \item \textbf{0.75}: Key decisions or interpersonal interactions with high value.
        \item \textbf{1.00}: Core system-generated memory points (fixed).
    \end{itemize}

    \vspace{2mm}
    \textbf{Generation Requirements} \\
    1. \textit{Atomic focus}: Each memory must record a single core fact without redundant modifiers. \\
    2. \textit{Temporal grounding}: Use standardized MMM DD, YYYY, hh:mm:ss format. \\
    3. \textit{Calibration}: Secondary memories must complement existing system memories without duplication, focusing on emotional nuances and specific thought processes.

    \vspace{2mm}
    \textbf{Output Format (JSON)}
\begin{lstlisting}[language=json, basicstyle=\tiny\ttfamily, breaklines=true, frame=single, backgroundcolor=\color{gray!10}]
{
  "start_time_point": "Jan 15, 2025, 14:30:00",
  "end_time_point": "Jan 15, 2025, 15:30:00",
  "dialogue_goal": "...",
  "dialogue_summary": "...",
  "memory_points": [
    {
      "index": 1,
      "memory_content": "...",
      "memory_type": "Persona Memory",
      "is_update": "False",
      "original_memories": [],
      "importance": 0.25
    }
  ]
}
\end{lstlisting}
\end{promptbox}

\subsection{Conflict Emotion Generation Template}
% 人类通常在短期的对话中，不容易出现冲突的情绪。然而，将对话的时间轴拉长之后，用户对于同一个谈论主题，可能会有截然不同的两种对立或者相反的情绪，这种具有冲突的情绪一般在短时间内难以生成，因此我们通过人工构造冲突情绪的方式，来评估记忆系统对冲突情感的检测和推理能力。一个具有直观的冲突情绪的例子如：会话1：“我很喜欢A基金，收益很稳定”；会话2（24h后）：“A基金最近暴跌，我现在很厌恶它了”；测试：“我现在对A基金的情感是什么？这个例子就要求记忆系统能检测到有冲突的情绪，并且在回答的时候给出正确的答案。

In typical human interactions, conflicting emotions rarely manifest within short-term dialogues. However, when the temporal scale of the conversation is extended, a user may exhibit diametrically opposed or contradictory affective states regarding the same subject matter. Since such emotional conflicts are difficult to generate spontaneously within a brief window, we employ a method of manual synthesis to construct conflicting emotional scenarios. This approach is designed to evaluate the memory system's capacity for conflict detection and affective reasoning. 

A representative conflict scenario involves a user expressing positive sentiment toward an asset (e.g., "I like Fund A for its stability") in an initial session, followed by a drastic shift to negative sentiment ("I now loathe Fund A after its collapse") twenty-four hours later. Evaluating the system via the query "What is my current sentiment regarding Fund A?" tests its ability to detect emotional incongruity across extended temporal scales. Success requires the memory system to resolve these conflicting affective states and prioritize the most recent context through temporal reasoning.

\begin{promptbox}{Template for Emotional Conflict Memory Generation}
    \textbf{Role Description} \\
    You are a professional memory point analysis expert. Your task is to generate exactly \texttt{\{conflict\_memory\_count\}} emotional conflict memory points based on a provided memory pool. Each entry must represent a clear shift or contradiction in the user's emotional stance, preferences, or attitudes relative to the original records.

    \vspace{2mm}
    \textbf{Task Description} \\
    Based on the provided original memory points, you must synthesize conflict memories that:
    \begin{itemize}
        \item Focus on the user's \textbf{emotional attitude, subjective preference, or value judgment}.
        \item Represent a clear \textbf{contradiction or reversal} of the original affective state.
        \item Remain \textbf{logically plausible} (e.g., mood swings or changes of heart) to challenge the agent's consistency.
    \end{itemize}

    \vspace{2mm}
    \textbf{Generation Requirements} \\
    1. \textit{Type Consistency}: Conflict points must maintain the same type as the source (Persona, Event, or Relationship). \\
    2. \textit{Emotional Flip}: Content must exhibit an explicit reversal, such as transitioning from enthusiasm to indifference or from support to opposition. \\
    3. \textit{Sensitivity Testing}: Focus on \textit{Attitude/Preference/Value/Mood Swing} to rigorously test the agent's sensitivity to emotional inconsistencies. \\
    4. \textit{Traceability}: Each generated point must map to a unique \texttt{source\_memory\_index} from the provided pool.

    \vspace{2mm}
    \textbf{Output Format (JSON)}
\begin{lstlisting}[language=json, basicstyle=\tiny\ttfamily, breaklines=true, frame=single, backgroundcolor=\color{gray!10}]
{
  "emotional_conflict_memories": [
    {
      "memory_content": "Conflict memory point content (reversed attitude)",
      "memory_type": "Persona / Event / Relationship Memory",
      "reason": "Reason for generating this emotional conflict",
      "source_memory_index": 1,
      "conflict_angle": "Attitude/Preference/Value/Emotional State"
    }
  ]
}
\end{lstlisting}

    \vspace{2mm}
    \textbf{Input Context} \\
    The generation is conditioned on the \texttt{original\_memory\_pool}, \texttt{memory\_type\_indices}, \texttt{persona\_info}, and the \texttt{existing\_memory\_list}.
\end{promptbox}

\subsection{Dialogue Generation Template}
% 结合前面的对话生成概要和冲突情绪信息生成。我们构建对话生成模板来创造对话。我们在构建对话的过程中，严格依赖于前面阶段构造的事件和时间线。利用LLM来生成自然流畅的多轮对话。在设计模板的过程中，我们还考虑到了测试维度。使得模板覆盖的信息更加完整。

Building upon the earlier dialogue abstracts and synthesized emotional conflicts, we designed a generation template to orchestrate multi-turn conversations. This process maintains strict fidelity to the pre-defined event sequences and temporal frameworks, utilizing LLMs to ensure linguistic fluidity. By embedding targeted evaluation dimensions within the template, we achieved a more holistic information architecture for testing memory systems.

\begin{promptbox}{Dialogue Generation Prompt Template}
    \textbf{Role Description} \\
    You are a professional dialogue generation expert tasked with synthesizing natural, multi-turn scripts between a user and an assistant. Your objective is to embed specific memory-testing dimensions—including real memory points and \textbf{emotional conflict memory points}—to evaluate the assistant's mnemonic resilience and profile-tracking capabilities.

    \vspace{2mm}
    \textbf{Core Testing Dimensions} \\
    The script must seamlessly integrate the following three adversarial metrics:
    \begin{itemize}
        \item \textit{Key Entity Extraction}: Capturing multiple attributes (time, location, actions) in a single turn.
        \item \textit{Long-Distance Anaphora}: Correctly resolving ambiguous references (e.g., "that meeting") across multiple intervening turns.
        \item \textit{Hallucination Inhibition}: Ensuring the assistant does not incorrectly record speculative or conflicting emotional stances as confirmed facts.
    \end{itemize}

    \vspace{2mm}
    \textbf{Emotional Conflict \& Interaction Logic} \\
    1. \textit{Assistant Behavior}: The assistant must confidently mention \textbf{emotional conflict memory points} (e.g., assuming a reversed preference) as if they are the user's current truth. \\
    2. \textit{User Behavior}: The user must \textbf{ignore} these inconsistencies, neither confirming nor correcting them, to test if the assistant erroneously updates the profile based on unverified statements. \\
    3. \textit{Response Depth}: Assistant replies must be 3--5 times longer than typical AI responses, providing deep analysis, actionable guidance, and rich contextual integration.

    \vspace{2mm}
    \textbf{Constraints \& Quality Metrics} \\
    \begin{itemize}
        \item \textit{Fixed Turns}: Generate exactly \texttt{\{turn\_num\}} turns, no more, no less.
        \item \textit{Persona Alignment}: User utterances must strictly adhere to \texttt{\{persona\_info\}} and \texttt{\{user\_motivation\_type\}}.
        \item \textit{Information Embodiment}: Every provided memory point must be explicitly or implicitly reflected in the dialogue flow.
    \end{itemize}

    \vspace{2mm}
    \textbf{Output Format (JSON)}
\begin{lstlisting}[language=json, basicstyle=\tiny\ttfamily, breaklines=true, frame=single, backgroundcolor=\color{gray!10}]
{
  "dialogue_turn_1": [
    {"role": "user", "content": "..."},
    {"role": "assistant", "content": "[Detailed response integrating conflicts]"}
  ],
  "dialogue_turn_2": [
    {"role": "user", "content": "[Ignoring emotional reversal]"},
    {"role": "assistant", "content": "..."}
  ]
}
\end{lstlisting}
\end{promptbox}

\subsection{Question Generation Template for Dataset Construction}

To rigorously evaluate the cognitive and emotional intelligence of long-term memory systems, we designed a multi-dimensional question generation template. By grounding the inquiry process in this comprehensive taxonomy, the template ensures that the evaluation transcends surface-level information retrieval to capture the sophisticated affective reasoning.

\begin{promptbox}{Question Generation Template}
    \textbf{Role Description} \\
    You are a professional Memory and Emotional Intelligence Evaluation Expert. Your task is to generate \texttt{\{question\_num\}} high-quality test questions for the \texttt{\{question\_type\}} category, strictly adhering to the HUMAINE EARL emotional standards and the provided user profile and dialogue history.

    \vspace{2mm}
    \textbf{Emotional Data Constraints} \\
    All generated emotional metadata must strictly follow the HUMAINE EARL taxonomy comprising 49 emotion types. 
    \begin{itemize}
        \item \textbf{Intensity Mapping}: Identify specific tags based on hierarchy (e.g., \textit{High}: rage; \textit{Medium}: angry; \textit{Low}: irritated).
        \item \textbf{Polarity}: Explicitly distinguish between \textit{positive} and \textit{negative} emotion types as defined in the background schema.
    \end{itemize}

    \vspace{2mm}
    \textbf{Generation \& Structural Requirements} \\
    1. \textit{Structural Alignment}: Answers for Recall tasks must follow the format: \texttt{[Emotion] ([Tag]) - [Reason]}. \\
    2. \textit{Temporal Precision}: Every question must include a temporal anchor (e.g., "In the 2025-Q2 Summary") to ensure grounding. \\
    3. \textit{Profile Interlinking}: For coherence questions, use static fields (e.g., \textit{Personality: Perfectionist}) to explain specific intensity threshold triggers. \\
    4. \textit{Evidence Integrity}: The \texttt{evidence} field must contain the exact raw string from the provided core memory points.

    \vspace{2mm}
    \textbf{Difficulty Grading Strategy} \\
    \begin{itemize}
        \item \textbf{Easy}: Direct matching of an explicit emotion tag within the text.
        \item \textbf{Medium}: Mapping descriptive dialogue or implicit cues to the correct Intensity Tag.
        \item \textbf{Hard}: Detecting irony, complex emotional reasoning, or resolving contradictory emotional types (Conflict Detection).
    \end{itemize}

    \vspace{2mm}
    \textbf{Output Format (JSON)}
\begin{lstlisting}[language=json, basicstyle=\tiny\ttfamily, breaklines=true, frame=single, backgroundcolor=\color{gray!10}]
{
  "questions": [
    {
      "question": "Regarding the [Event] on [Date], what was the user's emotion?",
      "answer": "Emotion: [name], Tag: [tag], Intensity: [Level]. Reason: [...]",
      "evidence": [{"memory_content": "...", "memory_type": "..."}],
      "difficulty": "hard"
    }
  ]
}
\end{lstlisting}
\end{promptbox}

% 其余的模板包含在压缩文件中。我们在压缩文件中添加了源代码、配置文件和说明信息。只需要按照说明进行详细的配置即可运行数据生成和系统评估代码。生成的数据和系统评估结果保存在独立的文件夹中。

Additional templates, along with source code and configuration files, are available in the attached archive. Following the provided instructions for environment setup and configuration will enable the automated data generation and evaluation suite. The resulting data and performance metrics are automatically saved in separate output folders.

\end{document}